\documentclass[journal]{IEEEtran}

\ifCLASSINFOpdf
\else
\fi

\usepackage{amsmath, amsthm, amssymb}
\usepackage{algpseudocode}
\usepackage{algorithm}
\usepackage{multirow}
\usepackage{array}
\usepackage[lofdepth,lotdepth]{subfig}
\usepackage{booktabs}
\usepackage{bm}
\usepackage{graphicx}
\usepackage{xcolor}
\usepackage{soul}
\usepackage[utf8]{inputenc}
\usepackage[figuresright]{rotating}
\usepackage{subfig}
\usepackage{subfiles}
\usepackage{float}
\usepackage{subfloat}
\usepackage{stfloats}
\usepackage{cases}
\usepackage{mdwmath}

\usepackage{tcolorbox}
\usepackage{authblk}

\usepackage[normalem]{ulem}
\usepackage{siunitx}
\usepackage{hyperref}

\begin{document}
 
\newtheorem{theorem}{Theorem}
\newtheorem{lem}{Lemma}
\newtheorem{cor}{Corollary}

\theoremstyle{definition}
\newtheorem{defn}[theorem]{Definition} %
\title{Multiscale Tensor Summation Factorization as a new Neural Network Layer (MTS Layer) for Multidimensional Data Processing}
\author[1]{Mehmet Yama\c{c}}
\author[1]{Muhammad Numan Yousaf}
\author[2]{Serkan Kiranyaz}
\author[1]{Moncef Gabbouj}
\affil[1]{Tampere University, Faculty of Information Technology and Communication Sciences, Tampere, Finland}
\affil[2]{Department of Electrical Engineering, Qatar University, Qatar}

\maketitle

\begin{abstract}
Multilayer perceptrons (MLP), or fully connected artificial neural networks, are known for performing vector-matrix multiplications using learnable weight matrices; however, their practical application in many machine learning tasks, especially in computer vision, can be limited due to the high dimensionality of input-output pairs at each layer. To improve efficiency, convolutional operators have been utilized to facilitate weight sharing and local connections, yet they are constrained by limited receptive fields. In this paper, we introduce Multiscale Tensor Summation (MTS) Factorization, a novel neural network operator that implements tensor summation at multiple scales, where each tensor to be summed is obtained through Tucker-decomposition-like mode products. Unlike other tensor decomposition methods in the literature
 MTS is not introduced as a network compression tool; instead, as a new backbone neural layer. MTS not only reduces the number of parameters required while enhancing the efficiency of weight optimization compared to traditional dense layers (i.e., unfactorized weight matrices in MLP layers), but it also demonstrates clear advantages over convolutional layers. The proof-of-the-concepts experimental comparison of the proposed MTS networks with MLPs and Convolutional Neural Networks (CNNs) demonstrates their effectiveness across various tasks, such as classification, compression, and signal restoration.  Additionally, when integrated with modern non-linear units such as the multi-head gate (MHG), also introduced in this study, the corresponding neural network, MTSNet, demonstrates a more favorable complexity-performance tradeoff compared to state-of-the-art transformers in various computer vision applications, including image restoration tasks such as denoising,  draining, and deblurring. The software implementation of the MTS layer and the corresponding MTS-based networks, MTSNet(s), is shared at \url{https://github.com/mehmetyamac/MTSNet}.

\end{abstract}


\IEEEpeerreviewmaketitle

\section{Introduction}
\label{Introduction}
\begin{figure}[t]
\centering
  \includegraphics[width=0.85\linewidth]{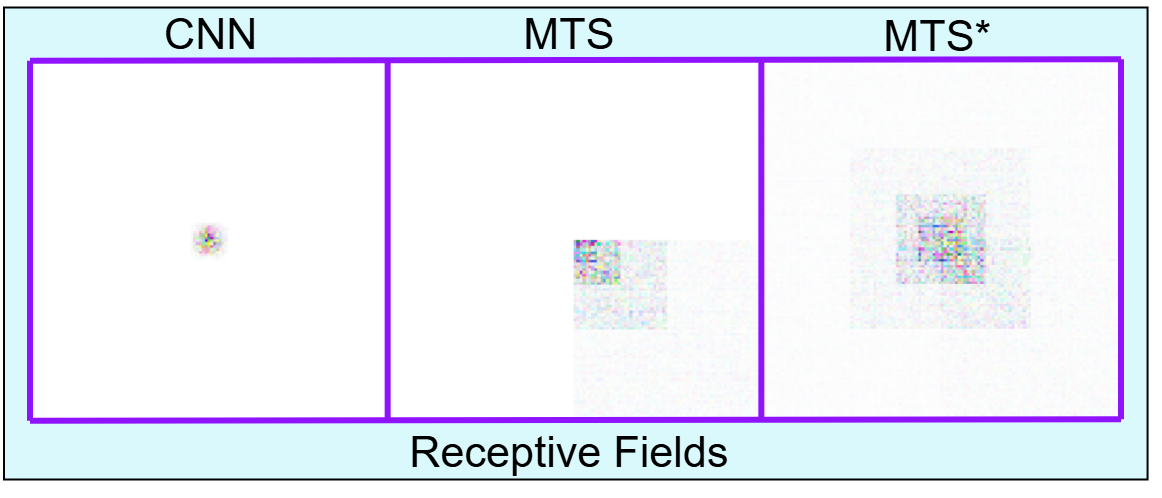}
 \caption{Comparison of receptive fields in vanilla networks for image restoration tasks: (a) convolutional neural network, (b) neural network with only MTS layers, (c) neural network from (b) with a convolution in the last layer.}

  \label{fig:ReceptiveField}
  \includegraphics[width=0.99\linewidth]{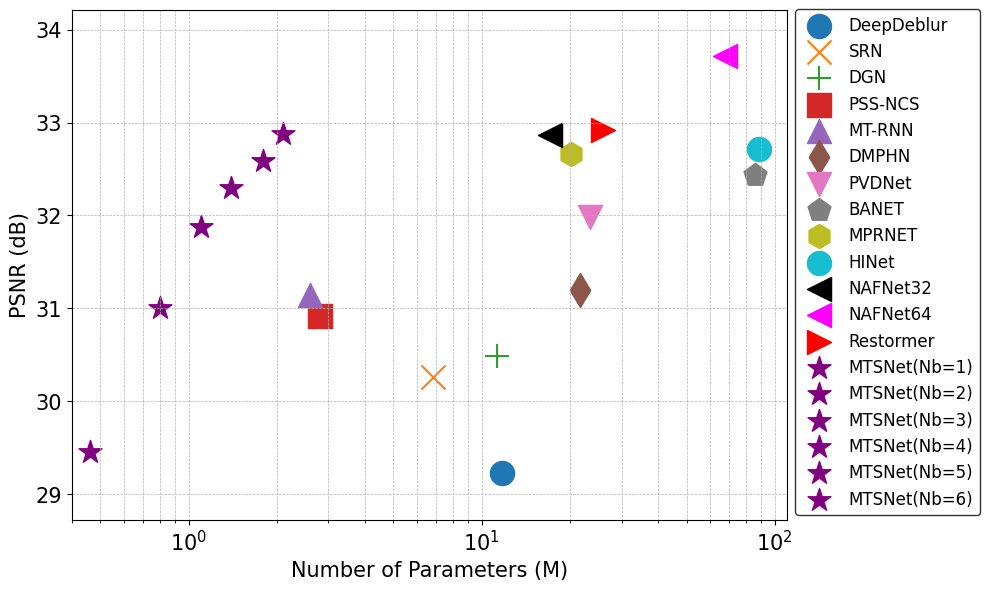}
  \caption{Comparison of existing deblurring networks in GoPro~\cite{gopro2017} Dataset.}
\label{fig:deblurring_all}
\end{figure}

\IEEEPARstart{M}ultilayer perceptrons (MLP) \cite{MLP1,MLP2}, or fully-connected Artificial Neural Networks, essentially perform vector-matrix multiplications using learnable weight matrices. However, learning such matrices is often impractical for numerous machine learning tasks, including computer vision, due to the infeasibility of making such a matrix-vector multiplication when considering multi-dimensional and large-size input-output signals at each layer. To address this, convolutional operators \cite{CNN1} have been introduced, which can be formalized as specific factorization for matrix-vector multiplication, creating an equivalent weight matrix that utilizes weight sharing and relies only on local connections. While convolutional layers in various network models have demonstrated remarkable performance over the years, they also come with certain limitations, notably their limited receptive field size due to the fixed kernel sizes. In this study, we re-examine the core question of whether an alternative matrix factorization method can improve representation capabilities. We introduce Multiscale Tensor Summation (MTS) Factorization, a novel neural network operator that applies tensor summation at multiple scales, where each core tensor is obtained via first mode product over multidimensional and then the summation of the predefined number, \(T\), of resulting tensors. 

In decomposing the full matrix-vector multiplication matrix, MTS can significantly reduce the number of parameters compared to the dense counterpart, and this helps to reduce the search space in learning the optimal (or sub-optimal) weight matrix (i.e., reducing the search space process with a minimal reduction in representation power). Such an efficient factorization makes the proposed MTS layer much more efficient than dense layers that learn unfactorized weight matrices. Although a conventional Tucker Decomposition, such as tensor rank reduction methods, can significantly reduce the number of parameters compared to the unfactorized version of the multiplication matrix, their usage in Artificial Neural Networks (ANNs) remains limited due to a significant decrease in representation power  \cite{yamac1}. The authors of \cite{yamac1, yamac2} introduced the idea of leveraging the summation of T number of mode products in factorizing the matrix instead of a single one in Tucker-Decomposition. Theoretical analysis is also given for linear compression (e.g., compressive sensing \cite{CS}) in terms of coherence metric \cite{cai2011limiting} (a metric used to measure the goodness of compression matrix in compressive sensing literature) in \cite{yamac2}. Nonetheless, their study remains restricted to tasks involving the learning of compressive sensing matrices and cannot be generalized as an alternative to dense layers, since the proposed GTS operation is applicable to only full images. In contrast, the MTS factorization introduced in this study offers a more flexible approach, utilizing a transformer-like patch division strategy \cite{ViT, ViT2}. Such a flexible factorization allows the operation to be applicable to large-size and multidimensional tensors, whether they are feature map tensors or input images. In addition, in the proposed MTS, patch division is conducted using various patch sizes (referred to as window sizes, which are essentially a list of predefined sizes for each look). This approach yields a more enriched feature representation. The feature map tensors generated by different scale-looks are then summed, resulting in a compact representation within the MTS layer.

The novel and significant contributions in this paper can be summarized as follows:

\begin{itemize}
    \item Introduction of a novel backbone neural operator, the MTS Layer, which serves as a superior alternative to conventional counterparts such as dense and convolutional layers.

    \item  The development of a new gate operation, the Multi-Head-Gate operation (MHG), which provides a better alternative to current non-linear activation functions such as ReLU.

    \item A novel non-linear layer that combines the MHG operator with MTS layer, proposed as an alternative to the existing backbones, particularly in comparison with state-of-the-art (SoTa) attention mechanisms, such as multi-head attention layers of transformers.

    \item Creation of the MTSNet toolbox, designed for seamless integration of MTS and MHG layers into existing neural architectures, with a current implementation available for PyTorch.
    
\end{itemize}

The direct comparison of MTS layers to common factorization techniques, such as convolution, reveals that MTS layers, first of all, demonstrate superiority thanks to their large and structured receptive fields that operate on different scales. Figure \ref{fig:ReceptiveField} illustrates how large receptive fields can be achieved even in small-scale vanilla networks. The significant gain of MTS layers over convolutional ones in terms of receptive field size is obvious as can be seen in this figure.   Second, when this linear operation is combined with modern non-linear units such as the proposed MHG operations, it can outperform transformers in various computer vision tasks. The comparative evaluations are performed in various experiments, including classification, compression, and various image restoration tasks, such as denoising and deblurring, for both compact and deep networks. The rest of the paper is organized as follows: Section \ref{related works} gives a summary of the literature on Tensor Decomposition methods developed for Deep Neural Networks (DNNs). Then, the proposed MTS layer is introduced in Section \ref{MTS_section}. In Section \ref{Results_section}
 we first compare the MTS layer with both the dense layer and SoTa tensor decomposing network TT-Net in single-layer image classification and compression tasks. Then, the MTS layer is compared with a conventional convolutional layer in image restoration tasks. Finally, a new non-linear unit, MHG, is introduced in an MTSNet layer that becomes a similar architecture to the SoTa transformer for image restoration, SwinIR \cite{swinir}, in various image restoration tasks. Finally, the conclusion and feature work discussions are drawn in Section \ref{conclusion}. 
\section{Related Works}
\label{related works}
In the literature, along with many studies on using tensor decomposition as an extension of conventional dimensional reduction techniques \cite{TT-SVD1, TT-SVD2} (e.g., SVD), numerous tensor decomposition techniques, have recently been proposed as an attempt to leverage tensor decomposing technology in DNNs; however, these models are primarily introduced as compression methods for DNN architectures \cite{TT-2015, TT-2017, TR-2019, tensorcompress1}. For instance, advanced tensor decomposition approaches, such as tensor train (TT) \cite{TT-2015} and tensor ring (TR) \cite{TR-2019}, have achieved a parameter count compression rate of \(O(1k)\) by factorizing the hidden layers of DNNs \cite{TT-2017}. This impressive compression rate has been attainable only in specific applications, particularly in the video compression of RNN-based networks. Nevertheless, the tensor rank reduction technique inherently results in information loss, thus restricting its applicability to broader problems. In image classification tasks, for example, even with just 10 classes, such as in the CIFAR-10 dataset, a notable information loss of up to \%2 has been observed, even with 6-fold compression \cite{li2021heuristic}. The techniques proposed in the literature for DNN compression can be categorized into two general groups \cite{yin2021towards}: (i) Learning a compressed network directly with hidden units of low tensor rank by training networks from scratch, pre-modelled as TT models, and (ii) Re-approximating each hidden layer of a network composed of pre-trained classical dense or convolutional layers using tensor decomposition techniques.

The ongoing research into tensor-based DNN compression methodologies continues to produce significant advancements \cite{yin2021towards}. For a comprehensive overview of current developments in this area, we refer interested readers to recent extensive surveys \cite{tensorsurvey}. On the other hand, we do not introduce the MTS factorization technique proposed in this study as a network compression method technique; on the contrary, we propose it as a backbone data representation unit, just like convolutional layers or self-attention layers.

MTSNets are designed to be trained in a multi-scale tensor summation format (i.e., from-scratch training strategy similar to the above-mentioned first type of tensor decomposition-based DNN compression approaches). This method differs from the conventional practice of initially training dense or convolutional layers and subsequently decomposing them into a tensor factorization format. \textit{Unlike other tensor factorization methods that adopt a similar strategy to ours, called from-scratch training strategy, which often leads to training instability (as noted in \cite{yin2021towards}), we demonstrate through experimental results and ablation studies that MTSNets maintain stability during training, whether in compact (shallow) networks or deep models. }


\section{MultiScale Tensorial Summation Factorization}
\label{MTS_section}
\begin{figure}[h]
\centering
  \includegraphics[width=0.92\linewidth]{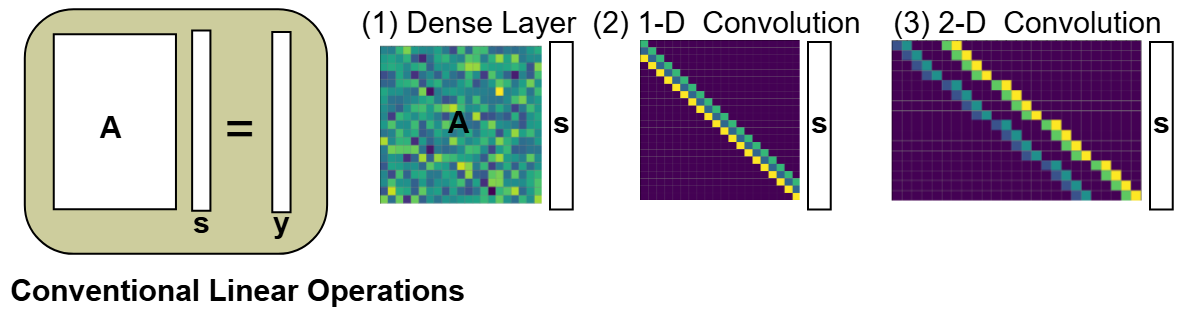}
  \caption{Conventional linear layers in matrix-vector multiplication form. }
\label{fig:linearlayers}
\end{figure}

Let an input image be represented as \(\mathbf{S} \in \mathbb{R}^{\sqrt{n} \times \sqrt{n}}\), for which we want to perform a linear operation. In conventional dense layer architectures, this image is initially flattened into a vector \(\mathbf{s} \in \mathbb{R}^{n \times 1}\). Subsequently, it is transformed by means of a learned matrix \( \mathbf{A} \in \mathbb{R}^{m \times n} \), leading to the formulation \(\mathbf{y} = \mathbf{A} \mathbf{s}\), where \(\mathbf{y} \in \mathbb{R}^{m \times 1}\) is interpreted as the feature map at the next layer. Figure \ref{fig:linearlayers} illustrates a pictorial representation of commonly used linear operators in vector-matrix multiplication form. 

If the transformation matrix can be expressed as a Kronecker product of distinct left and right matrices, it yields a separable linear operation, defined as:

\begin{equation}
    \mathbf{Y} = \mathbf{A_1} \mathbf{S} \mathbf{A_2}, \label{seperable1}
\end{equation}

where \(\mathbf{A_1} \in \mathbb{R}^{\sqrt{m} \times \sqrt{n}}\) and \(\mathbf{A_2} \in \mathbb{R}^{\sqrt{m} \times \sqrt{n}}\) represent the left and right transformation matrices. A notable instance of this can be observed in the conventional two-dimensional Discrete Cosine Transform (DCT) operation, which can be executed using a one-dimensional DCT matrix \(\mathbf{A_1}\) and its transpose \(\mathbf{A_2}\), applied to the image \(\mathbf{S}\) from both the left and right sides, respectively.

Given the formulation in Eq. \eqref{seperable1}, it is possible to revert to the conventional matrix-vector multiplication form:

\begin{equation}
     \mathbf{y} = \text{vec}(\mathbf{Y}) = (\mathbf{A_1} \otimes \mathbf{A_2})~ \text{vec}(\mathbf{S})  = \mathbf{A} \mathbf{s} ,
\end{equation}
where \(\otimes\) indicates the Kronecker product operation. Rather than employing a conventional non-learnable matrix pair, one can train a neural network to learn the coefficients for the left and right matrices \cite{yamac1}. This approach results in a significant reduction in the number of trainable parameters, decreasing them from \( m \times n \) to \( 2 \times \sqrt{m} \times \sqrt{n} \) in comparison to the dense layer configuration that appears in standard MPLs. Additionally, in terms of computational complexity, the count of multiplication operations (Multiply-Accumulate, MAC) required for the linear transformation is also diminished, reducing the calculations from \( m \times n \) to \(  2 \times \sqrt{m} \times n \).

In a typical ANN, we are interested in applying the transformation not only to matrices but also to multidimensional signals, i.e., both large-size and multidimensional input data and feature maps in each hidden layer. Let \( \mathbf{\mathcal{S}} \in \mathbb{R}^{n_1 \times n_2 \times \ldots \times n_J} \) be the signal of interest, which is also called the feature tensor at a specific layer. The next layer feature map can be obtained via a multi-linear transformation,
\begin{equation}
    \mathbf{\mathcal{Y}} = \mathbf{\mathcal{S}} \times_1 \mathbf{A_1} \times_2 \mathbf{A_2} \ldots \times_J \mathbf{A_J}, \label{seperable2}
\end{equation}
where \( \mathbf{\mathcal{S}} \times_i \mathbf{A_i} \) is the \( i \)-mode product applied on the tensor \( \mathbf{\mathcal{S}} \) and the matrix \( \mathbf{A_i} \in \mathbb{R}^{m_i \times n_i} \) is the learnable multiplication matrix along \(i^{th}\) dimension, and \( \mathbf{\mathcal{Y}} \in \mathbb{R}^{m_1 \times m_2 \times \ldots \times m_J}  \) is the output tensor of the layer, which corresponding the feature map at the layer in a typical compact or DNN with total \( m =\prod_{j=1}^J m_j \) number of coefficients. Similar to the 2-D case, the multidimensional linear operation can also be cast as a standard linear one,
\begin{equation}
    \text{vec}( \mathbf{\mathcal{Y}} )= \left ( \mathbf{A_1} \otimes \mathbf{A_2} \otimes \ldots \otimes \mathbf{A_J} \right ) \text{vec}( \mathbf{\mathcal{S}}).
\end{equation}
\subsection{Factorizing dense linear matrix as the sum of Kronecker linear transformation matrices}
In spite of the fact that it may be possible to reduce the memory and computation requirements by using the Tucker-Decomposition style learnable factorization defined in Eq. \eqref{seperable2}, such a separable linear system may cause significant information loss compared to using standard dense layer which learn the multiplication matrix in an unfactorized manner \cite{yamac1}. In order to compensate for such an information loss, the authors of \cite{yamac1} and \cite{yamac2} introduced the summed factorization,
\begin{equation}
    \mathbf{\mathcal{Y}} = \sum_{t=1}^{T} \mathbf{\mathcal{S}} \times_1 \mathbf{A_1^{(t)}} \times_2 \mathbf{A_2^{(t)}} \ldots \times_{J-1} \mathbf{A_{J-1}^{(t)}} \times_J \mathbf{A_J^{(t)}}, \label{tensorsum1}
\end{equation}
where \(T\) is the number of different separable output feature map tensors and \( \mathbf{A_i^{(t)}} \) denotes the \( i^{th} \) dimension multiplication matrix for the \( t^{th} \) operation. Let the Kronecker Product of \( t^{th} \)-term be \( \mathbf{{P}}^{t} = \mathbf{A_1^{(t)}} \otimes \mathbf{A_2^{(t)}} \otimes \ldots \otimes \mathbf{A_J^{(t)}}  \). The corresponding equivalent linear operation can be written in the form  \( \mathbf{y} = \mathbf{{P}}^{1} \mathbf{s} + \mathbf{{P}}^{2} \mathbf{s} + ...+ \mathbf{{P}}^{T} \mathbf{s}  = (\mathbf{{P}}^{1} + \mathbf{{P}}^{2} ... +\mathbf{{P}}^{T}   ) \mathbf{s}  \). Thus, the factorized multiplication matrix \(\mathbf{P}\) can be written
\begin{equation}
    \mathbf{{P}} = \sum_{t=1}^{T}  \left ( \mathbf{A_1^{(t)}} \otimes \mathbf{A_2^{(t)}} \otimes \ldots \otimes \mathbf{A_J^{(t)}} \right ), \label{tensorsum2}
\end{equation}
when we write the vectorized version of measurements, i.e., \( \mathbf{y} = \mathbf{P} \mathbf{s} \). It is important to note that while one can compute \( \mathbf{P} \) and substitute it for any conventional dense layer, in practical applications, this is neither necessary nor efficient, as it results in a complexity equivalent to that of using a dense layer. This system reduces the number of trainable parameters from \( \prod_{j=1}^{J} m_j n_j \) to \( T \sum_{j=1}^{J} m_j n_j \) compared to unfactorized counterpart in a typical dense layer.
\subsection{Multi-Scale Tensorial Summation Factorization}
\begin{figure*}[]
\centering
  \includegraphics[width=0.65\linewidth]{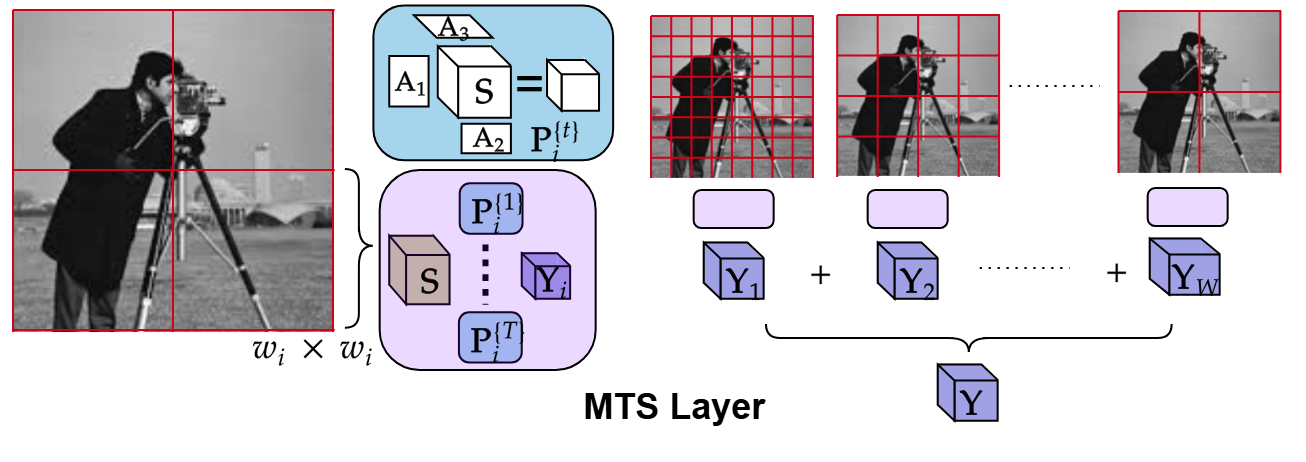}
  \vspace{-0.2cm}
  \caption{A MTS layer operates at different scales to obtain feature map \(\mathbf{\mathcal{Y}}\). }
\label{fig:MTSLayer}
\end{figure*}
\begin{figure}[]
\centering
  \includegraphics[width=0.8\linewidth]{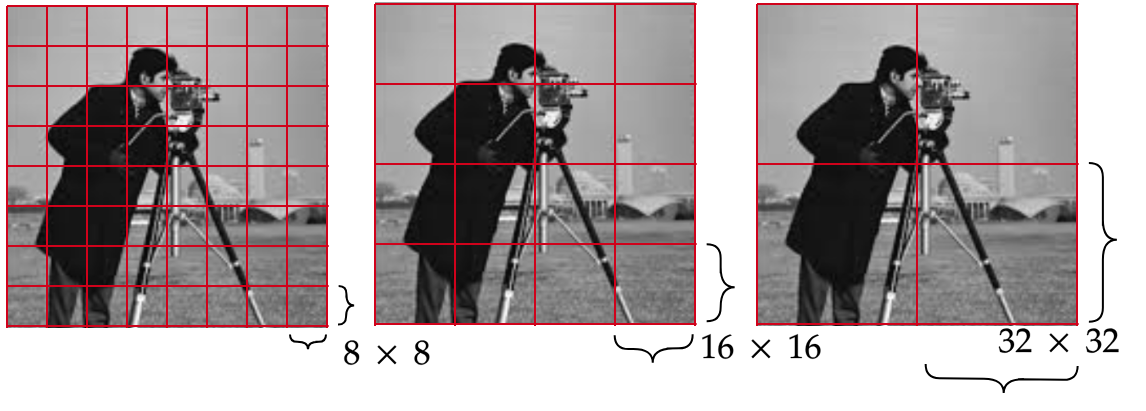}
  \vspace{-0.2cm}
  \caption{MTS Layer applies tensorial summation factorization in different scales. In a transformer, linear operation is applied patch-wise at each hidden layer (before non-linear operation) but for fixed sizes of patches. In MTS, multilinear tensorial factorization is applied on different-sized patches, and then the result is further summed.  }
\label{fig:windows}
\end{figure}

The system with learnable parameters described in Eq. \eqref{tensorsum2} was introduced in \cite{yamac1}, specifically for application to the input image in order to learn the compressive sensing matrix. However, its implementation within the hidden layers of any DNN architecture remains impractical. To enhance its feasibility, we propose using this operation in a patch-wise manner rather than applying it to the entire feature tensor. Additionally, conventional factorization techniques, such as convolution, are effective tools since they exhibit shift-invariance. However, they are not scale-invariant; the receptive field size is constrained by the kernel size, and increasing the kernel size leads to a quadratic increase in network complexity. As an alternative, we suggest applying the operation defined in Eq. \eqref{tensorsum2} at multiple scales by utilizing different patch sizes (window sizes), followed by a summation of the output feature map tensors. A graphical illustration of this concept can be found in Figure \ref{fig:windows}. Patch-wise operations have gained popularity in recent computer vision applications, particularly in transformers \cite{ViT, ViT2, swinir}. In contrast to the proposed multi-scale tensorial summation (MTS) layer, typical transformers operate using fixed-sized windows. Mathematically speaking, let \( \mathbf{\mathcal{X}} \in \mathbb{R}^{H \times W \times C} \) be our feature map to be multi-linearly transformed. We define the  \(w_i \times w_i\)-size patch embedding function as \(f_{w_i}: \mathbb{R}^{H \times W \times C} \to \mathbb{R}^{B_i \times w_i \times w_i \times C}\), where \(B_i\) is the number of patches for the corresponding scale \(i\). The linear operation is applied in a multi-linear manner to each of them, independently. In contrast to the linear operations in transformer-based neural network architectures, where such a patch-wise operation is typically performed by flattening, i.e., \( f^{\text{transformers}}_{w_i}: \mathbb{R}^{H \times W \times C} \to \mathbb{R}^{B_i \times D} \), where \(D =  w_i^2 C \) is the dimension of each flattened patch, our straightforward patch operation, \(f_{w_i}\), preserves spatial structure of the signal. 

In an MTS layer, a patch-wise multilinear transformation is applied for various patch sizes according to a predefined list, denoted as \(w = [w_1, w_2, ...w_{SC}]\), where \(SC\) represents the number of scales. Then, the resulting feature tensors are summed to have the final output of the MTS layer. In essence, MTS is an extension of GTS (Generalized Tensorial Sum Operation in Eq. \eqref{tensorsum1}), which is suitable for multi-dimensional signals (or latent space signals, e.g., feature maps, etc.),
\begin{equation}
\begin{aligned}
\hspace{-0.2cm} \mathcal{Y} = \sum_{sc=1}^{SC} f^{-1}_{w_{sc}} \Bigg( \sum_{t=1}^{T} & f_{w_{sc}}(\mathcal{X}) 
\times_1 \mathbf{A}_1^{(t,sc)} 
\times_2 \mathbf{A}_2^{(t,sc)}  
\times_3 \dots \\
& \times_{J-1} \mathbf{A}_{J-1}^{(t,sc)} 
\times_J \mathbf{A}_J^{(t,sc)} \Bigg), \!
\end{aligned} \label{MTS}
\end{equation}
where \(f^{-1}_{w_{sc}}\left (  \right )\) is the inverse patch-embedding operation of the corresponding window size, \(w_{sc}\). In here we denote $SC$ as the corresponding scale (e.g., for window size \( w\)= $[8]$, $SC=1$ means GTS that is applied $8\times8$ block wise divided $\mathcal{S}$, on the other hand, if \(w=[8,16]\), it means that \(SC=2\) and GTS is applied both \(8 \times 8\) patches and \(16 \times 16 \) ones, and the corresponding feature maps is obtained by summing the result of this two operations. In the equation, \(T\) is the number of different separable tensors, and 
\(
\mathbf{A}_j^{(t, i)} \in \mathbb{R}^{(w_{i,\text{out}})^{(j)} \times w_i^{(j)}}
\) represents the transformation matrix along the \(j^{\text{th}}\) dimension for the \(t^{\text{th}}\) operation at scale \(i\), where \(w_i^{(j)}\) and \((w_{i,\text{out}})^{(j)}\) denote the input and output window sizes, respectively, for the corresponding scale.
In the matrix-vector multiplication form of such a linear operation, the corresponding linear transformation matrix can be obtained via
\begin{equation}
\hspace{-0.1cm} \mathbf{P} = \sum_{sc=1}^{SC} I_{B_{sc}} \otimes \sum_{t=1}^{T}  \left( \mathbf{A}_1^{(t,sc)} \otimes \mathbf{A}_2^{(t,sc)} \otimes \dots \otimes \mathbf{A}_J^{(t,sc)} \right),
\end{equation}
where each term \(I_{B_{sc}} \otimes \sum_{t=1}^{T} \left( \mathbf{A}_1^{(t,sc)} \otimes \dots \otimes \mathbf{A}_J^{(t,sc)} \right)\)  represents a block-diagonal matrix corresponding to a patch-wise linear operator. It is constructed as a sum of \(\prod_{j=1}^{J} (w_{sc,\text{out}}^{(j)}) \times \prod_{j=1}^{J} w_{sc}^{(j)}\) matrices obtained via Kronecker multiplication. Therefore, the multi-linear operation can be equivalently expressed in the conventional matrix-vector multiplication form. However, in practical applications, particularly when dealing with large-scale signals such as images or feature maps, this approach is often neither necessary nor feasible. The operation defined in Eq. \eqref{MTS} corresponds to the aforementioned MTS layers, with learnable multiplication matrices, \(\mathbf{A}_j^{(t,sc)}\). A pictorial representation of the proposed MTS layer is shown in Figure \ref{fig:MTSLayer}. Throughout this study, we assume \(J = 3\), with the third mode corresponding to the channel dimension, i.e., \(w_{sc}^{(3)} = C\), and equal window sizes for the first two modes, i.e., \(w_{sc}^{(1)} = w_{sc}^{(2)}\).
\section{Example Network Designs and Experimental Evaluations}
\label{Results_section}
\subsection{Dense Layer vs MTS Layer}
\label{denselayervsmts}
In this section, we will compare the proposed MTS Layer with the standard dense layer counterpart. In the following section, preliminary experiments will be conducted on two different tasks: image classification and compression. 

\subsubsection{Image Classification}
In this prof-of-the-concept experimental design, one hidden layer that is fully connected (standard MLP) for image classification will be considered. Such a dense layer is also factorized by the MTS layer or by one of the SoTa tensor decomposition-based NN classification techniques, TT-Net \cite{TT-2015, TTNet2}.  A ReLU activation function is employed, followed by a hidden layer. The experiments are conducted using the MNIST, CIFAR-10~\cite{Krizhevsky2009-cifar}, STL-10 \cite{STL} datasets \cite{krizhevsky2009learning}, where the image sizes are \( 32 \times 32 \) for MNIST,  \( 3 \times 32 \times 32 \) for CIFAR-10 and \( 3 \times 96 \times 96 \) for STL-10. Even in this simplified example, when a multilayer perceptron (MLP) is utilized, a single layer necessitates \( 1024 \times 1024 \) parameters for MNIST and \( 3072 \times 3072 \) parameters for CIFAR-10. In contrast, when employing tensor factorizations such as TT-Net or the MTS layer, there is a significant reduction in the number of parameters. Fully connected last layers for each case can not be implemented in a patch-wise manner, but can still be decomposed by either MTS (reduced to be GTS layer) or TT-Net.  
\begin{figure}[H]
\centering
  \includegraphics[width=0.49\linewidth]{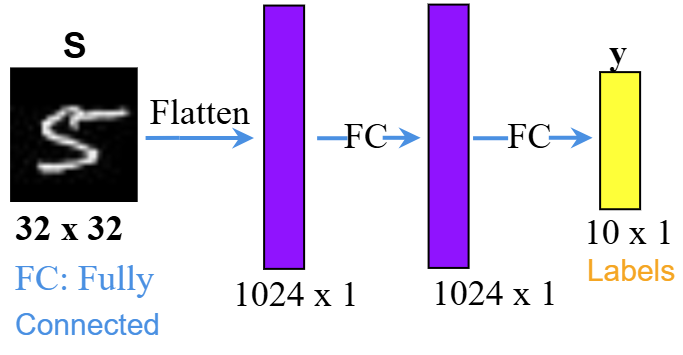}
  \includegraphics[width=0.49\linewidth]{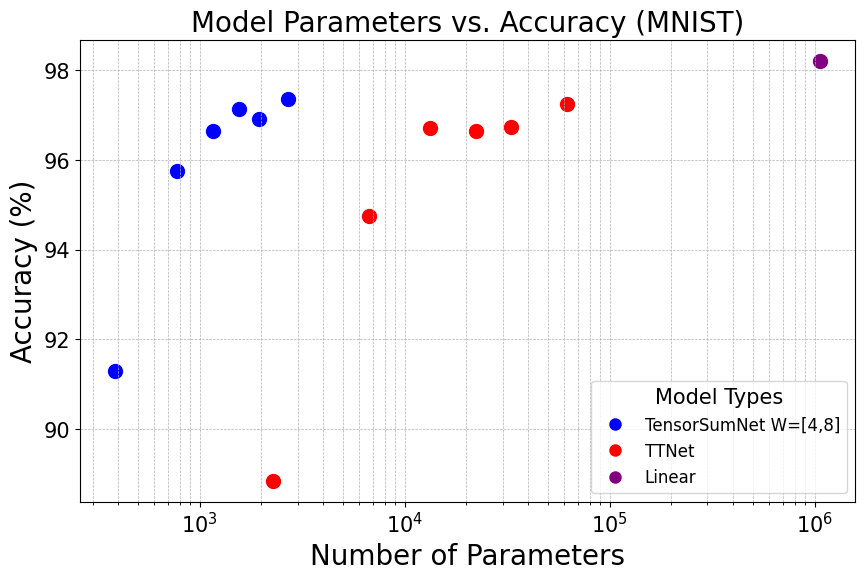}
  \includegraphics[width=0.50\linewidth]{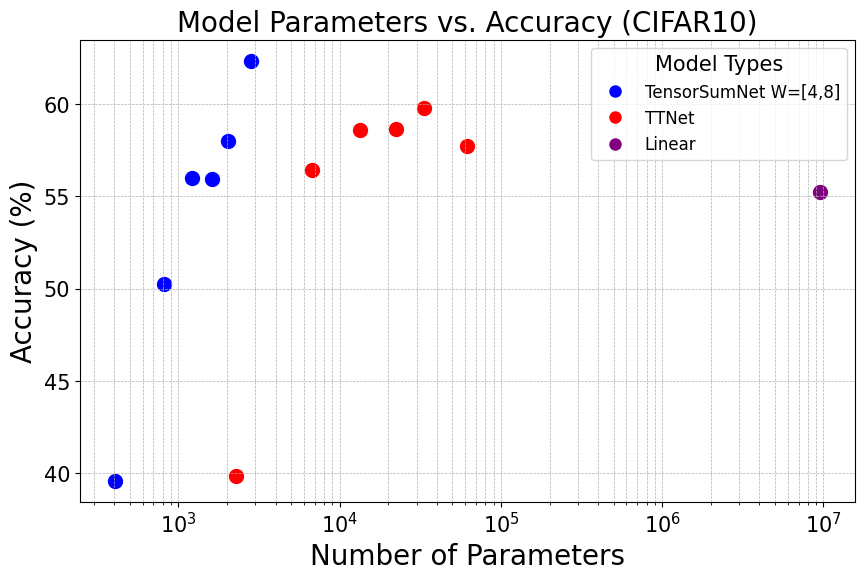}
  \hspace{-0.25cm}
  \includegraphics[width=0.50\linewidth]{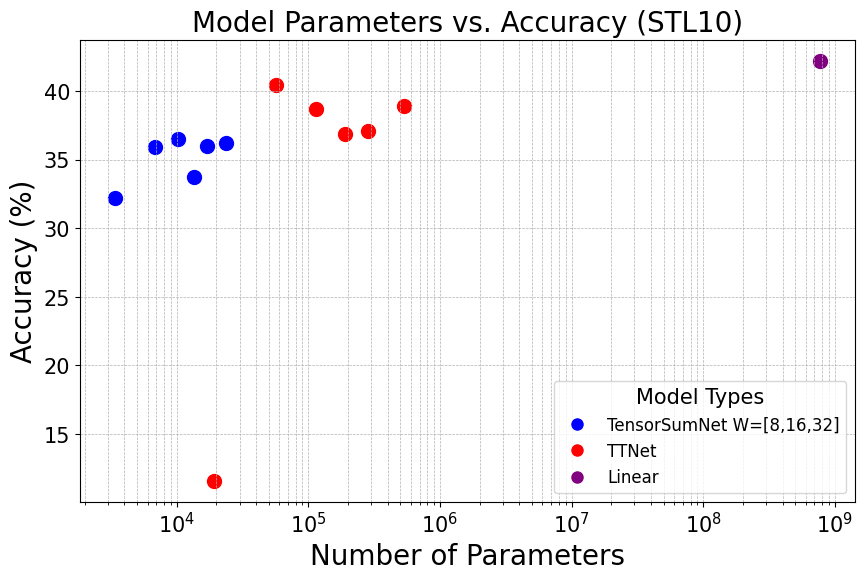}
  \caption{Proof-of-the-concept image classification using a single hidden layer classifier. A total of 3 layers are either implemented by standard dense layers, or they are factorized by two competing tensor decomposition methods, TT-Net and TensorSum (which use MTS Layers). }
\label{fig:classification model}
\end{figure}

As depicted in Figure \ref{fig:classification model}, experiments were conducted utilizing varying values of \( T \) for the corresponding MTS layer network,  i.e., TensorSumNet, alongside differing rank parameters for competing TTNet. The results indicate that the MTS layer maintains considerable representational capacity while achieving a significant reduction in the number of trainable parameters compared to standard MLP or SoTa tensorial factorization, TTNet. For network training, we used the SGD with momentum optimizer \cite{goodfellow2016deep, ramezani2024generalization} with a momentum value of 0.99. The learning rate was set to \( lr = 1\text{e}^{-3} \) for the CIFAR-10 and MNIST datasets and \( lr = 1\text{e}^{-4} \) for the STL-10 dataset. A MultiStepLR scheduler was applied with decay milestones at [50, 75, 100, 125] epochs and a decay factor of \( \gamma = 0.5 \).  On the other hand, for the dense layer configuration, the learning rate had to be reduced to \( lr = 1\text{e}^{-4} \) to avoid instability in MLP layers at higher learning rates (i.e., \( lr = 1\text{e}^{-3} \)).   
\subsubsection{Image Compression}
\begin{figure}[h]
\centering
  \includegraphics[width=0.5\linewidth]{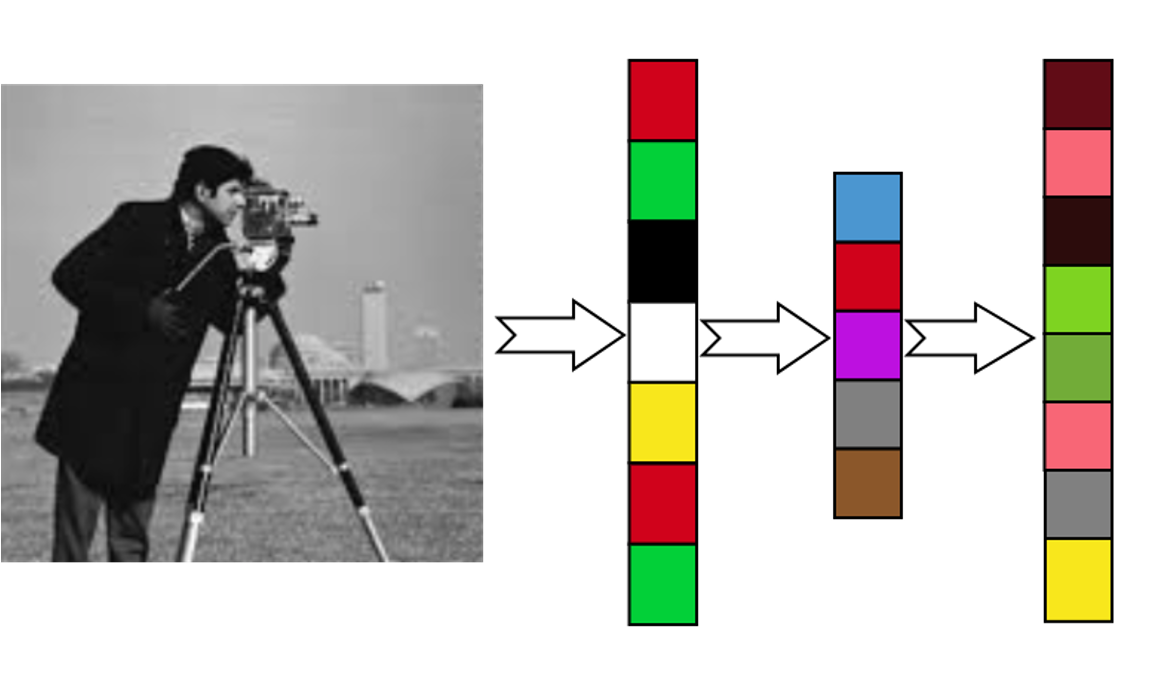}
  \hspace{-0.25cm}
  \includegraphics[width=0.5\linewidth]{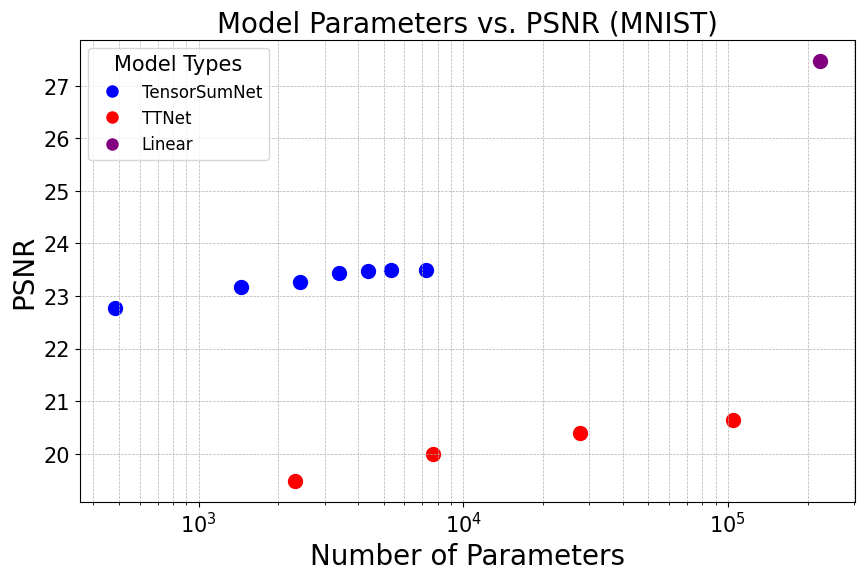}
  \hspace{-0.25cm}
  \includegraphics[width=0.5\linewidth]{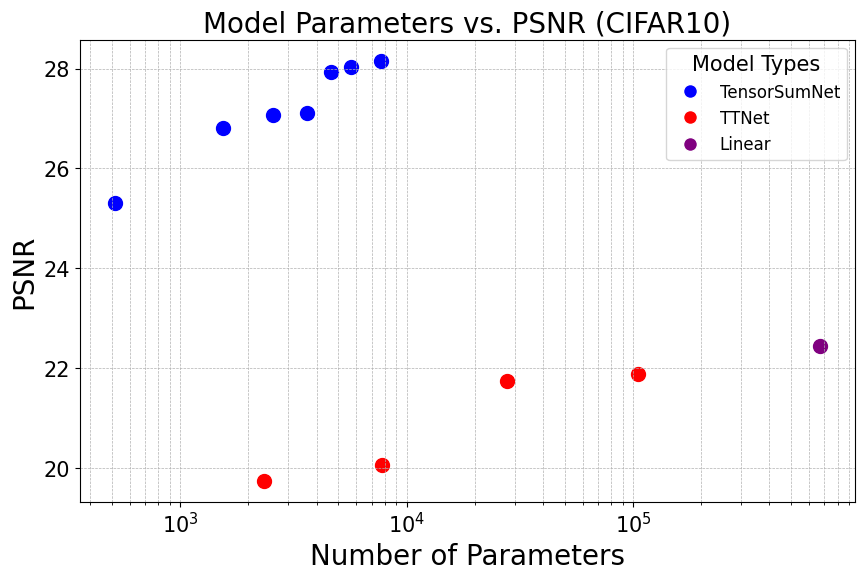}
  \hspace{-0.25cm}
  \includegraphics[width=0.5\linewidth]{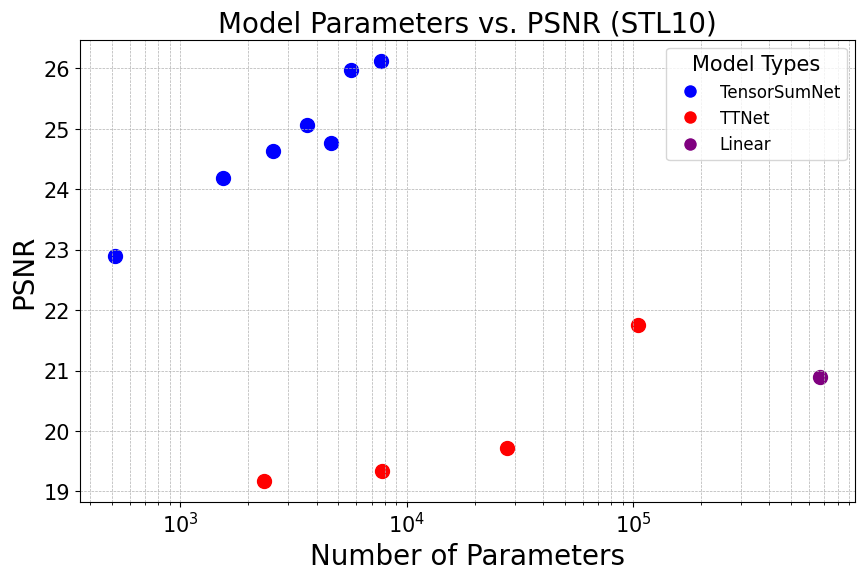}
  \caption{Prof-of-the-concept linear image compression model. A total of 2 layers are used. One is for compression, and the other is for decompression. In competing methods, either dense layers are used or factorized by TT-Net.}
\label{fig:compression_models}
\end{figure}

In this setup, a linear compression-uncompression scheme will be considered. Such linear embedding appears in many forms on all kinds of signal processing tasks, e.g., from dimensional reduction and compressive sensing \cite{tensorcompress1} to any type of linear feature extraction in DNN models, e.g., in transformers. Again a fully connected compression/uncompression model is either factorized by TT-Net or MTSNet. The input signal is reduced by a compression ratio of 1/8. Training was conducted on the MNIST, CIFAR-10, and STL-10 datasets. The Adam optimizer \cite{kingma2014adam} was used with learning rate,  \( lr = 1\text{e}^{-4} \), which decayed at epochs 50, 70, and 90 with a factor of 0.2 over 100 epochs. Models were trained using MSE loss, and the batch size was set to 16. Peak Signal-to-Noise Ratio (PSNR) was used as the evaluation metric during testing. As shown in Figure \ref{fig:compression_models}, the proposed MTS layer-based compression technique surpasses the standard MLP and TTNet by a large margin in terms of PSNR versus the number of learnable parameters.
\begin{figure}[h]
\centering
\includegraphics[width=0.75\linewidth]{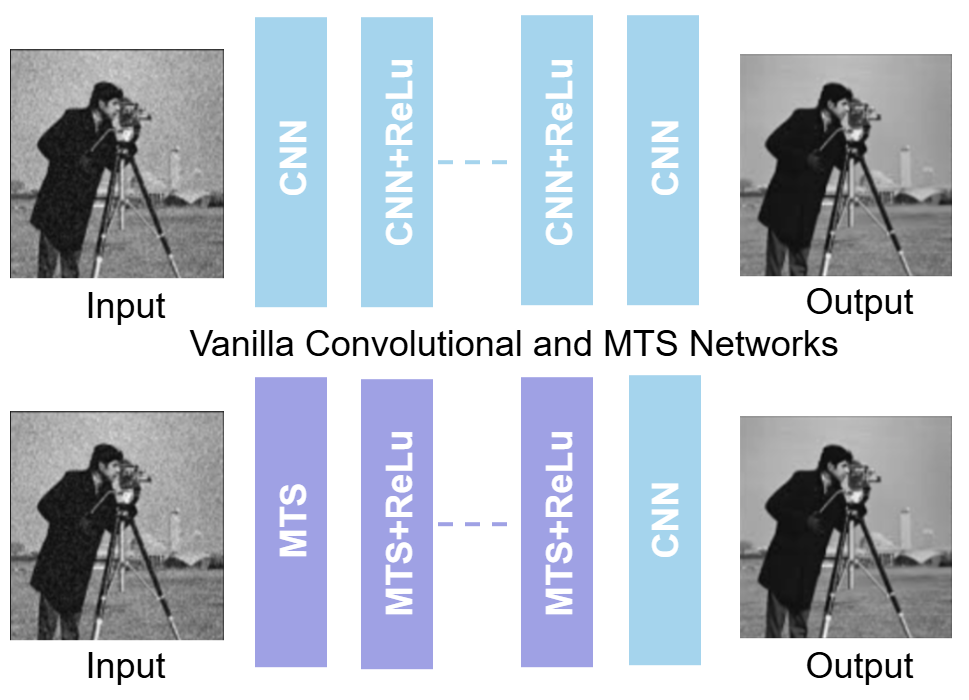}
\hspace{-0.55cm}
\includegraphics[width=0.35\linewidth]{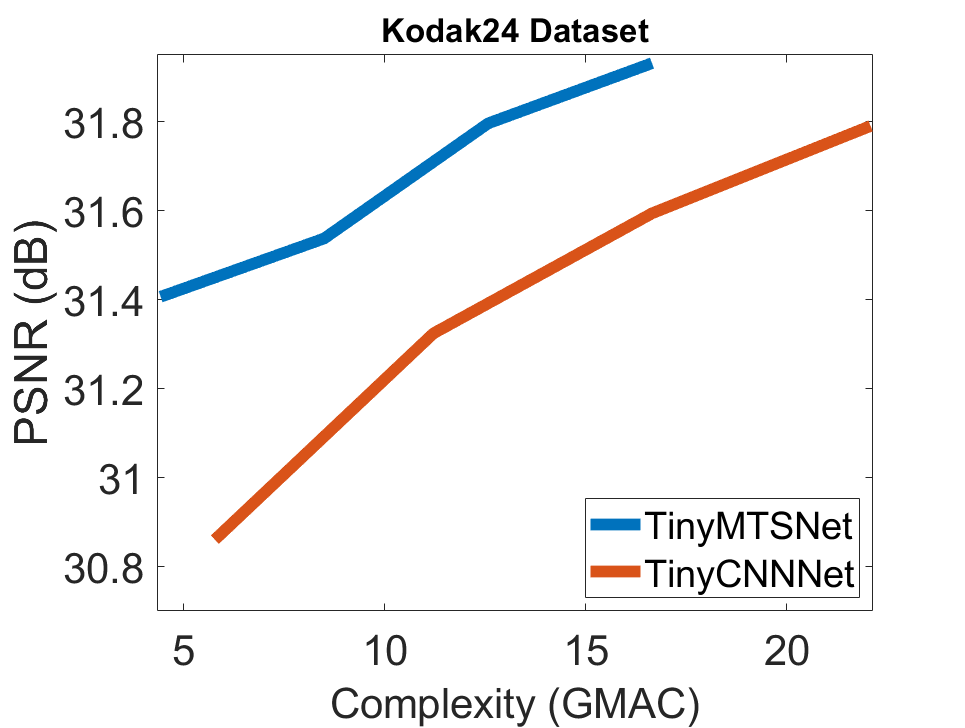}
\hspace{-0.55cm}
\includegraphics[width=0.35\linewidth]{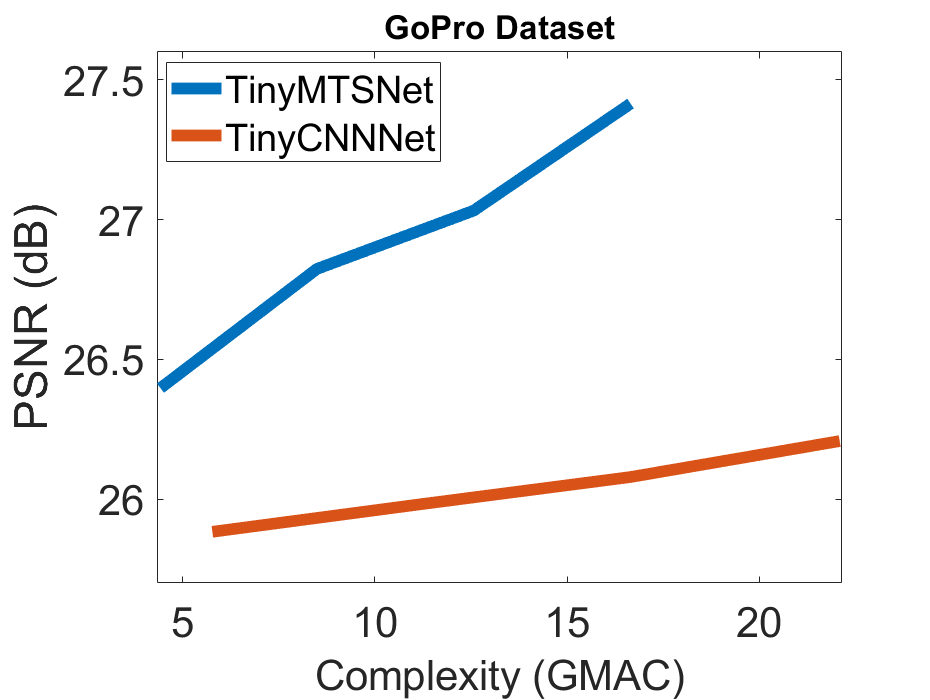}
\hspace{-0.55cm}
\includegraphics[width=0.35\linewidth]{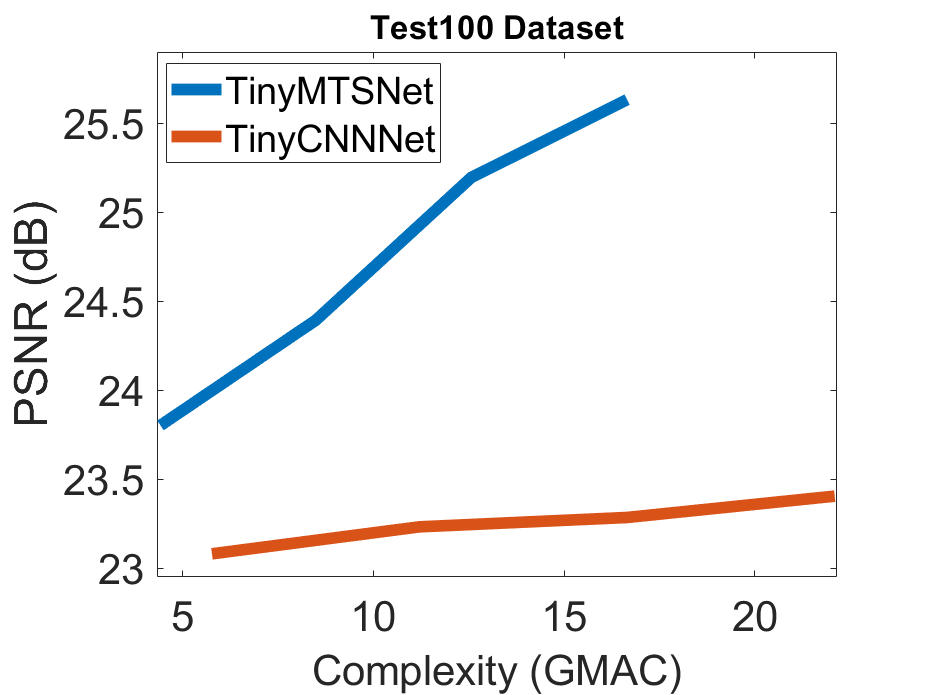}
\hspace{-0.99cm}
\includegraphics[width=0.35\linewidth]{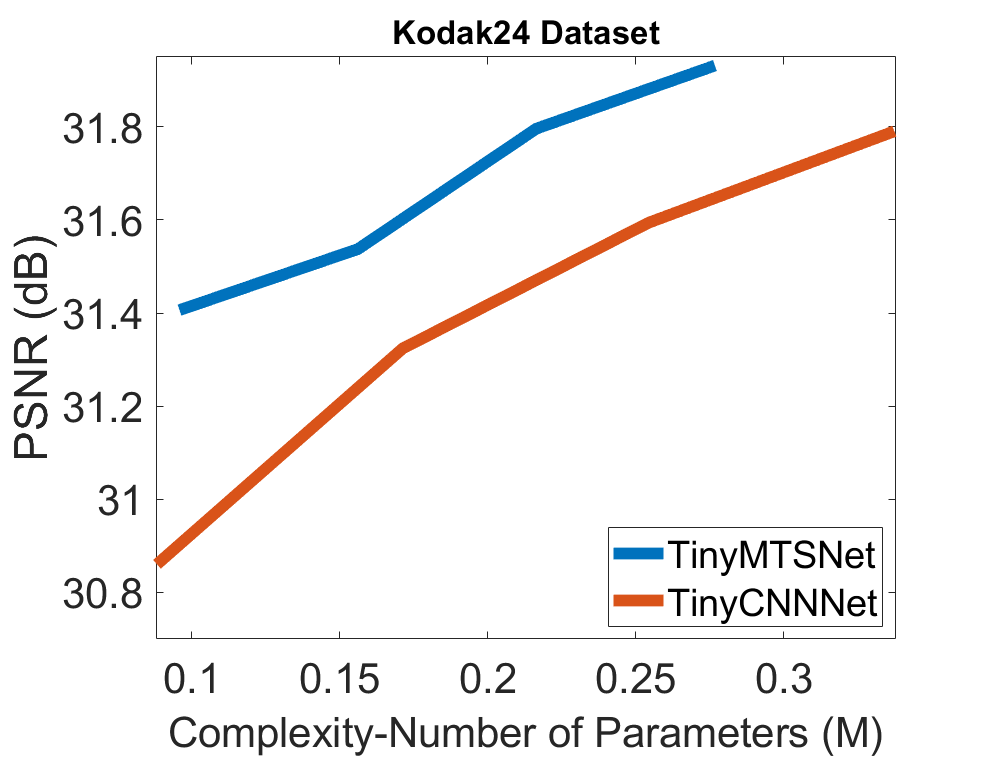}
\hspace{-0.5cm}
\includegraphics[width=0.34\linewidth]{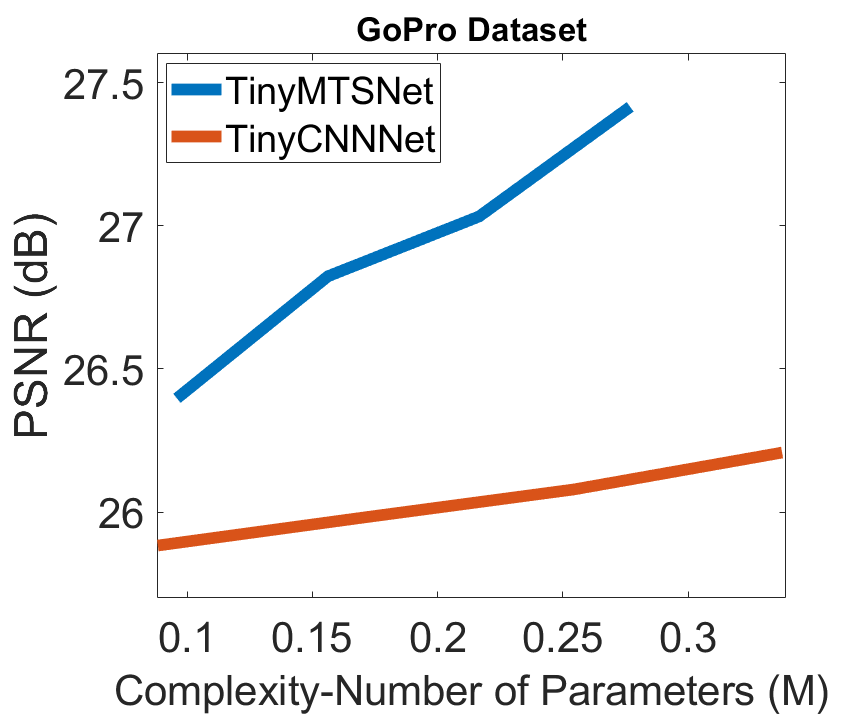}
\hspace{-0.38cm}
\includegraphics[width=0.35\linewidth]{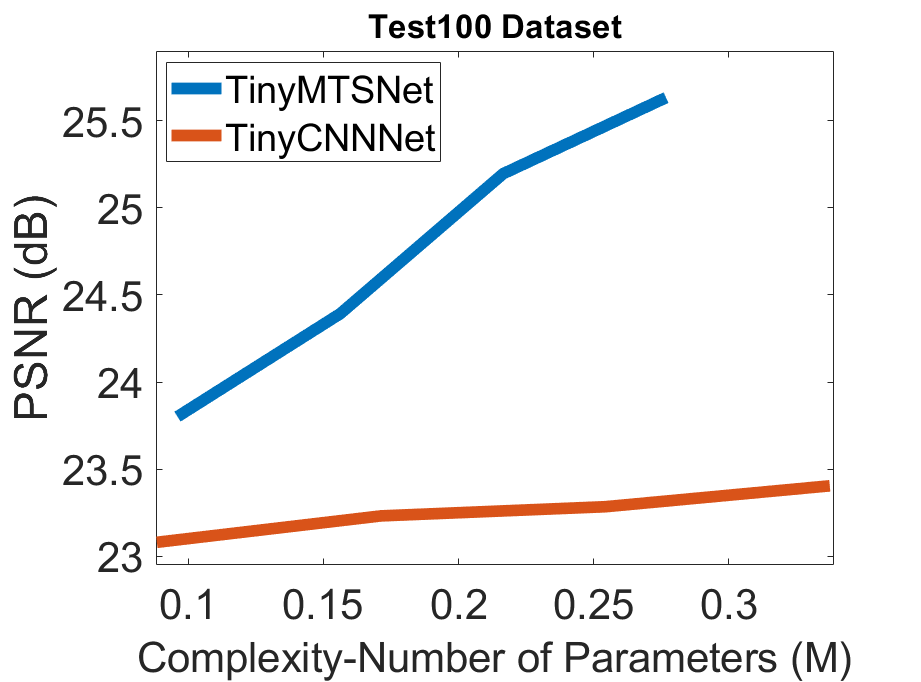}
  \caption{Architecture of vanilla networks, demonstrating two configurations: (i) convolutional layers, and (ii) MTS layers. Both networks contain a residual connection from the input to the output image and a ReLU activation at hidden layers. MTS Networks demonstrates its superiority compared to corresponding CNNs, in various image restoration tasks: image denoising (in Kodak24) dataset, image deblurring (in GoPro Dataset), and image deraining (in Test100). }
\label{fig:vanillamodels}
\end{figure}

\subsection{Convolutional Layer vs MTS Layer}
\label{convolutionvsMTS}
Convolutional is a well-established version of a specific factorization of the dense layer as illustrated in Figure \ref{fig:linearlayers}. We will compare MTS and CNN vanilla networks illustrated in Figure \ref{fig:vanillamodels} on image restoration tasks, denoising, deburring, and deraining. 
Both neural networks are configured with varying depths, ranging from one to four hidden layers. Each network has a residual connection from input to output. The datasets employed for training image denoising models include the Waterloo Exploration Database (WED) \cite{ma2016gmad}, DIV2K \cite{agustsson2017ntire}, BSD500~\cite{BSD500} and Flickr2K \cite{lim2017enhanced}, considering noise level of \(\sigma = 25\).  For the deblurring task, the GoPro~\cite{gopro2017} Dataset is used for both training and test cases. For deraining, Rain13K~\cite{MPRNET} is used in training networks. Patch size is set to \( 256 \times 256\) for all training setups, and batch size is configured to 32 for both deblurring and denosing, while it is adjusted to 16 for denoising tasks. The parameters are optimized by minimizing the L1 loss with AdamW~\cite{AdamW} optimizer ($\beta_1 = 0.9$, $\beta_2 = 0.999$, weight decay = 0, with an EMA decay rate of 0.9999) for all the models. MultiStepLR scheduler was applied with starting learning rate, \( lr = 1\text{e}^{-3} \), and decay milestones at [150k, 300k, 450k] iterations with a total of 600k iterations and a decay factor of $\gamma = 0.1$.     

\begin{figure}[]
\centering
  \vspace{-0.2cm}
  \includegraphics[width=0.95\linewidth]{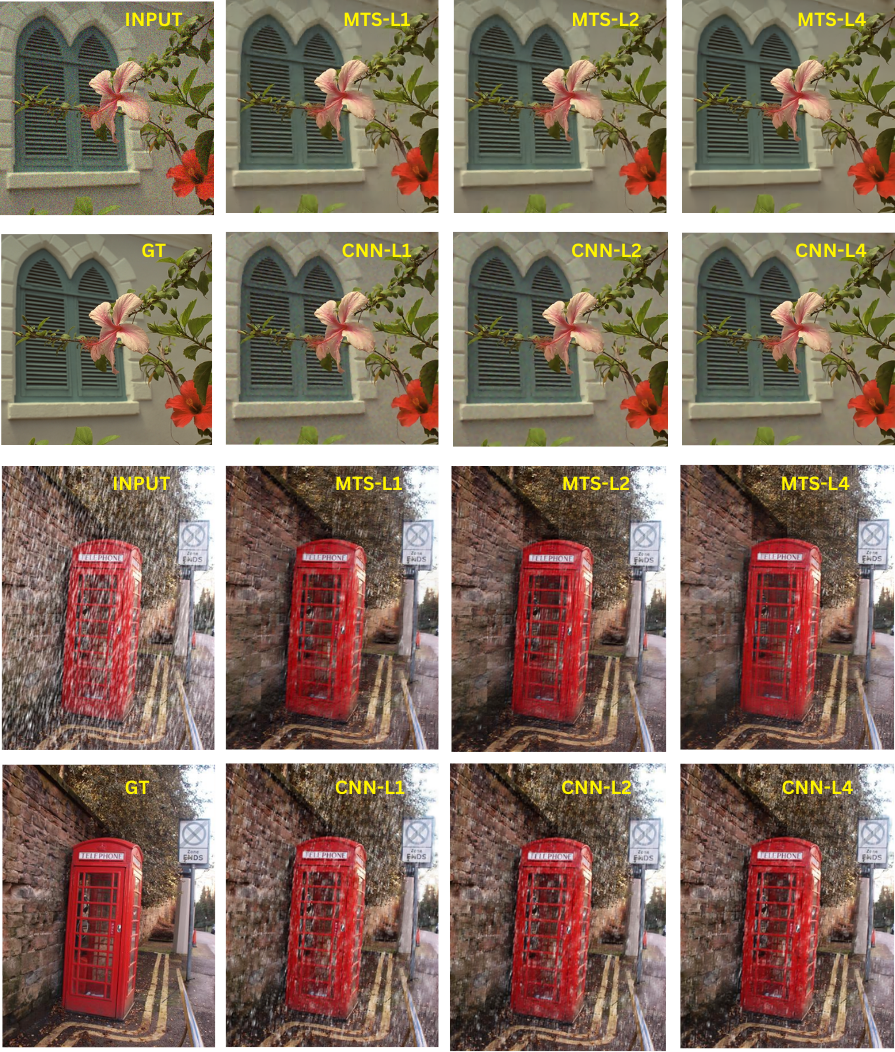}

  \caption{Comparative analysis on the visual quality of vanilla networks having MTS versus convolutional layers across various hidden layer numbers on different tasks and datasets, where \(L_i\) refers to using \(i\) number of hidden layers. }
\label{fig:MTS-CNN-visual}
\end{figure}

In order to keep the number of parameters and required MAC operations similar for both networks, the number of feature maps were set to
 \(C=96\) and \(C=48\) for CNNs and MTS Networks, respectively. In this preliminary experimental validation scheme, the MTS layer demonstrates its superiority as a backbone layer in the benchmark RGB image denoising on dataset Kodak24~\cite{kodak}, image deblurring dataset GoPro, and image de-raining dataset Test100~\cite{zhang2019image}, as shown in Figure~\ref{fig:vanillamodels}, thanks to its larger receptive field, and high data representation capacity. In Figure~\ref{fig:MTS-CNN-visual}, the first row presents an image denoising example, while the second row illustrates image deraining. For the denoising task, CNN-based vanilla networks exhibit noticeable noise artifacts, particularly the ones with a small number of hidden layers. In contrast, MTS-based networks produce significantly sharper outputs, even with just a single hidden layer. Improvements are particularly noticeable when looking closely at fine details (a zoom-in on the images reveals this). For image deraining, it is apparent that vanilla CNNs fail to effectively remove rain spots, whereas MTS-based networks produce satisfactory results, even the ones with only one hidden layer.

\subsection{Vision Transformer vs MTSNet}
\begin{figure}[h]
\centering
  \includegraphics[width=0.92\linewidth]{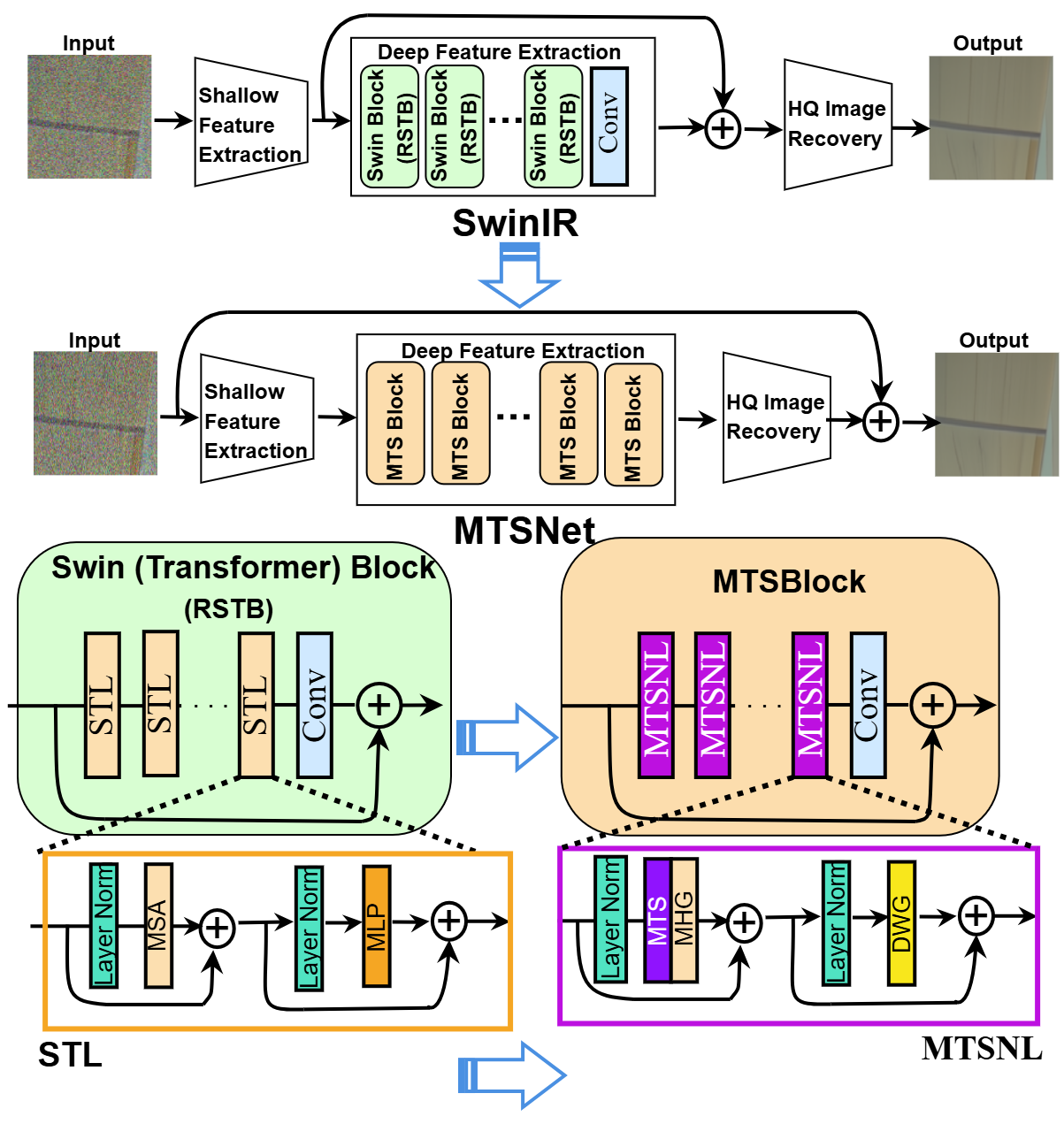}
  \caption{For the fair comparison,  we modify the SwinIR architecture by replacing its standard blocks (RSTBs) with MTSBlocks.  }
\label{fig:MTSBlock}
\end{figure}
Transformers can be considered backbone neural architectures with sub-modules that serve as alternatives to aforementioned conventional neural processing units, i.e., convolutions and dense layers. Although the Transformer~\cite{vaswani2017attention} model was originally introduced for natural language processing, transformer models have gained considerable attention in almost all fields of signal processing, including computer vision. For example, vision transformers were successfully applied to image classification~\cite{jia2024center, zhang2024catnet, ViT, liu2021swin, vaswani2021scaling}, detection~\cite{carion2020DETR, touvron2021training, su2023towards, fang2024eva, hou2024relation}, segmentation~\cite{xie2021segformer, xu2023levit, cao2022swin, cheng2022masked}, and so on. Among other applications, Transformer has also been introduced for image restoration~\cite{wang2022uformer, chen2021IPT, swinir, restormer} successfully.

\begin{table*}[] \scriptsize
\caption{The denoising results of color images with Gaussian noise were investigated on four test datasets and three different noise levels.}
\centering
\setlength\tabcolsep{2.6pt}
\begin{tabular}{lcccccccccccccc}
\hline
\multicolumn{1}{|l|}{Method}       & \multicolumn{3}{c|}{CBSD68~\cite{martin2001database_bsd}}                                                                                     & \multicolumn{3}{c|}{Kodak24~\cite{kodak}}                                                                                    & \multicolumn{3}{c|}{McMaster~\cite{zhang2011color_mcmaster}}                                                                                   & \multicolumn{3}{c|}{Urban100~\cite{huang2015single_urban100}}                                                  & GMACs                & \multicolumn{1}{c|}{\#Params (M)} \\
\multicolumn{1}{|l|}{}             & $\sigma=15$                  & $\sigma=25$                  & \multicolumn{1}{c|}{$\sigma=50$}                  & $\sigma=15$                  & $\sigma=25$                  & \multicolumn{1}{c|}{$\sigma=50$}                  & $\sigma=15$                  & $\sigma=25$                  & \multicolumn{1}{c|}{$\sigma=50$}                  & $\sigma=15$          & $\sigma=25$          & \multicolumn{1}{c|}{$\sigma=50$} &                      & \multicolumn{1}{c|}{}             \\ \hline
\multicolumn{1}{|l|}{IRCNN~\cite{zhang2017learning} }        & 33.86                        & 31.16                        & \multicolumn{1}{c|}{27.86}                        & 34.69                        & 32.18                        & \multicolumn{1}{c|}{28.93}                        & 34.58                        & 32.18                        & \multicolumn{1}{c|}{28.91}                        & 33.78                & 31.20                & \multicolumn{1}{c|}{27.70}       & -                    & \multicolumn{1}{c|}{0.18}         \\
\multicolumn{1}{|l|}{FFDNet ~\cite{FFDNetPlus}}       & 33.87                        & 31.21                        & \multicolumn{1}{c|}{27.96}                        & 34.63                        & 32.13                        & \multicolumn{1}{c|}{28.98}                        & 34.66                        & 32.35                        & \multicolumn{1}{c|}{29.18}                        & 33.83                & 31.40                & \multicolumn{1}{c|}{28.05}       & -                    & \multicolumn{1}{c|}{0.85}         \\
\multicolumn{1}{|l|}{DnCNN ~\cite{DnCNN} }        & 33.90                        & 31.24                        & \multicolumn{1}{c|}{27.95}                        & 34.60                        & 32.14                        & \multicolumn{1}{c|}{28.95}                        & 33.45                        & 31.52                        & \multicolumn{1}{c|}{28.62}                        & 32.98                & 30.81                & \multicolumn{1}{c|}{27.59}       & 37                   & \multicolumn{1}{c|}{0.56}         \\
\multicolumn{1}{|l|}{DSNet~\cite{peng2019dilated}}        & 33.91                        & 31.28                        & \multicolumn{1}{c|}{28.05}                        & 34.62                        & 32.16                        & \multicolumn{1}{c|}{29.05}                        & 34.67                        & 32.40                        & \multicolumn{1}{c|}{29.28}                        & -                    & -                    & \multicolumn{1}{c|}{-}           & -                    & \multicolumn{1}{c|}{-}            \\
\multicolumn{1}{|l|}{DRUNet~\cite{zhang2021DPIR}}       & 34.30                        & 31.69                        & \multicolumn{1}{c|}{28.51}                        & 35.31                        & 32.34                        & \multicolumn{1}{c|}{29.25}                        & 35.40                        & 33.14                        & \multicolumn{1}{c|}{30.08}                        & 34.81                & 32.60                & \multicolumn{1}{c|}{29.61}       & 144                  & \multicolumn{1}{c|}{32.64}        \\
\multicolumn{1}{|l|}{RPCNN~\cite{xia2020rpcnn}}        & -                            & 31.24                        & \multicolumn{1}{c|}{28.06}                        & -                            & 32.34                        & \multicolumn{1}{c|}{29.25}                        & -                            & -                            & \multicolumn{1}{c|}{-}                            & -                    & -                    & \multicolumn{1}{c|}{-}           & -                    & \multicolumn{1}{c|}{-}            \\
\multicolumn{1}{|l|}{BRDNet~\cite{tian2020BRDnet}}       & 34.10                        & 31.43                        & \multicolumn{1}{c|}{28.26}                        & 34.88                        & 32.41                        & \multicolumn{1}{c|}{29.29}                        & 35.08                        & 32.75                        & \multicolumn{1}{c|}{29.53}                        & 34.42                & 31.99                & \multicolumn{1}{c|}{28.56}       & -                    & \multicolumn{1}{c|}{-}            \\
\multicolumn{1}{|l|}{RNAN~\cite{zhang2019residual}}         & -                            & -                            & \multicolumn{1}{c|}{28.27}                        & -                            & -                            & \multicolumn{1}{c|}{29.58}                        & -                            & -                            & \multicolumn{1}{c|}{29.72}                        & -                    & -                    & \multicolumn{1}{c|}{29.08}       & 496                  & \multicolumn{1}{c|}{8.96}         \\
\multicolumn{1}{|l|}{RDN~\cite{zhang2020rdn}}          & -                            & -                            & \multicolumn{1}{c|}{28.31}                        & -                            & -                            & \multicolumn{1}{c|}{-}                            & -                            & -                            & \multicolumn{1}{c|}{-}                            & -                    & -                    & \multicolumn{1}{c|}{29.71}       & 1400                 & \multicolumn{1}{c|}{-}            \\
\multicolumn{1}{|l|}{IPT~\cite{chen2021IPT}}          & -                            & -                            & \multicolumn{1}{c|}{-}                            & -                            & -                            & \multicolumn{1}{c|}{-}                            & -                            & -                            & \multicolumn{1}{c|}{-}                            & -                    & -                    & \multicolumn{1}{c|}{29.98}       & 512                  & \multicolumn{1}{c|}{115.3}        \\
\multicolumn{1}{|l|}{SwinIR~\cite{swinir}}       & 34.42                        & 31.78                        & \multicolumn{1}{c|}{28.56}                        & 35.34                        & 32.89                        & \multicolumn{1}{c|}{29.79}                        & 35.61                        & 33.20                        & \multicolumn{1}{c|}{30.23}                        & 35.13                & 32.90                & \multicolumn{1}{c|}{29.72}       & 759                  & \multicolumn{1}{c|}{11.75}        \\
\multicolumn{1}{|l|}{Restormer~\cite{restormer}}    & 34.40                        & 31.79                        & \multicolumn{1}{c|}{28.60}                        & 35.47                        & 33.03                        & \multicolumn{1}{c|}{30.04}                        & 35.61                        & 33.34                        & \multicolumn{1}{c|}{30.37}                        & 35.15                & 32.96                & \multicolumn{1}{c|}{30.04}       & 141                  & \multicolumn{1}{c|}{26.13}        \\
\multicolumn{1}{|l|}{KBNet~\cite{zhang2023kbnet}}        & 34.41                        & 31.80                        & \multicolumn{1}{c|}{28.62}                        & 35.46                        & 33.05                        & \multicolumn{1}{c|}{30.04}                        & 35.56                        & 33.32                        & \multicolumn{1}{c|}{30.37}                        & 35.15                & 32.96                & \multicolumn{1}{c|}{30.04}       & 69                   & \multicolumn{1}{c|}{118}          \\ \hline
\multicolumn{1}{|l|}{MTSNet (NB1)} & 34.19                        & 31.56                        & \multicolumn{1}{c|}{28.36}                        & 34.97                        & 32.50                        & \multicolumn{1}{c|}{29.40}                        & 35.10                        & 32.77                        & \multicolumn{1}{c|}{29.66}                        & 34.44                & 32.01                & \multicolumn{1}{c|}{28.75}       & 26.36                & \multicolumn{1}{c|}{0.46}         \\
\multicolumn{1}{|l|}{MTSNet (NB2)} & {\color[HTML]{000000} 34.29} & {\color[HTML]{000000} 31.67} & \multicolumn{1}{c|}{{\color[HTML]{000000} 28.48}} & {\color[HTML]{000000} 35.10} & {\color[HTML]{000000} 32.65} & \multicolumn{1}{c|}{{\color[HTML]{000000} 29.57}} & {\color[HTML]{000000} 35.29} & {\color[HTML]{000000} 32.97} & \multicolumn{1}{c|}{{\color[HTML]{000000} 29.87}} & 34.7                 & 32.37                & \multicolumn{1}{c|}{29.16}       & 51.3                 & \multicolumn{1}{c|}{0.80}         \\
\multicolumn{1}{|l|}{MTSNet (NB3)} & {\color[HTML]{000000} 34.34} & {\color[HTML]{000000} 31.73} & \multicolumn{1}{c|}{{\color[HTML]{000000} 28.54}} & {\color[HTML]{000000} 35.18} & {\color[HTML]{000000} 32.73} & \multicolumn{1}{c|}{{\color[HTML]{000000} 29.65}} & {\color[HTML]{000000} 35.38} & {\color[HTML]{000000} 33.07} & \multicolumn{1}{c|}{{\color[HTML]{000000} 29.97}} & 34.84                & 32.53                & \multicolumn{1}{c|}{29.40}       & 76.23                & \multicolumn{1}{c|}{1.14}         \\
\multicolumn{1}{|l|}{MTSNet (NB4)} & {\color[HTML]{000000} 34.36} & {\color[HTML]{000000} 31.75} & \multicolumn{1}{c|}{{\color[HTML]{000000} 28.57}} & {\color[HTML]{000000} 35.22} & {\color[HTML]{000000} 32.77} & \multicolumn{1}{c|}{{\color[HTML]{000000} 29.68}} & {\color[HTML]{000000} 35.42} & {\color[HTML]{000000} 33.12} & \multicolumn{1}{c|}{{\color[HTML]{000000} 30.02}} & 34.91                & 32.62                & \multicolumn{1}{c|}{29.53}       & 101.16               & \multicolumn{1}{c|}{1.49}         \\
\multicolumn{1}{|l|}{MTSNet (NB5)} & {\color[HTML]{000000} 34.38} & {\color[HTML]{000000} 31.77} & \multicolumn{1}{c|}{{\color[HTML]{000000} 28.58}} & {\color[HTML]{000000} 35.24} & {\color[HTML]{000000} 32.79} & \multicolumn{1}{c|}{{\color[HTML]{000000} 29.71}} & {\color[HTML]{000000} 35.45} & {\color[HTML]{000000} 33.15} & \multicolumn{1}{c|}{{\color[HTML]{000000} 30.06}} & 34.96                & 32.70                & \multicolumn{1}{c|}{29.63}       & 126                  & \multicolumn{1}{c|}{1.83}         \\ \hline
                                   & \multicolumn{1}{l}{}         & \multicolumn{1}{l}{}         & \multicolumn{1}{l}{}                              & \multicolumn{1}{l}{}         & \multicolumn{1}{l}{}         & \multicolumn{1}{l}{}                              & \multicolumn{1}{l}{}         & \multicolumn{1}{l}{}         & \multicolumn{1}{l}{}                              & \multicolumn{1}{l}{} & \multicolumn{1}{l}{} & \multicolumn{1}{l}{}             & \multicolumn{1}{l}{} & \multicolumn{1}{l}{}             
\end{tabular}
\label{Table-I}
\end{table*}

A typical transformer network is composed of self-attention and feedforward modules arranged sequentially. In this study, we modify SwinIR, which is among the SoTa algorithms for image restoration, and conduct a comparative analysis. There are three modules in SwinIR: shallow feature extraction, deep feature extraction, and high-quality image reconstruction modules. We use the same network structure with SwinIR except in the deep feature extraction module, and MTS layer-based building blocks, MTSBlocks, are used instead of SwinIR's {R}esidual {S}win {T}ransformer {B}locks (RSTBs). Additionally, in the shallow feature extraction module, we include a parallel MTS layer alongside the convolution layer, differing from the original SwinIR, which only employs convolution. The same network architectures are applied for all image restoration tasks.  Figure \ref{fig:MTSBlock} illustrates the SwinIR network architecture, where SwinIR attention-based building blocks are replaced with MTS layer-based ones in order to have a fair comparison. 

In the shallow feature extraction module, \(F_{\textit{SF}}\), MTSNet takes a low-quality (LQ) input, \(\mathbf{\mathcal{S}}_{\textit{LQ}} \in \mathbb{R}^{H\times W\times 3}\), and produces a feature map \(\mathbf{\mathcal{X}}_0 \in \mathbb{R}^{H\times W\times C}\) via
\begin{equation}
    \mathbf{\mathcal{X}}_0 = F_{\textit{SF}}(\mathbf{\mathcal{S}}_{\textit{LQ}})
\end{equation}
where, \(C\) is the number of channels. The shallow feature extraction is done by fusing MTS and convolution layer feature extraction, i.e.,
\begin{equation}
    \mathbf{\mathcal{X}}_0 = \text{Conv}_{1 \times 1}\left( \text{Concat}\left( \text{Conv}_{3 \times 3}(\mathbf{\mathcal{S}}_{\textit{LQ}}),\ \text{MTS}(\mathbf{\mathcal{S}}_{\textit{LQ}}) \right) \right),
\end{equation}
where MTS layer has window size, \( w = [8,\ 16,\ 32,\ 64] \), and \(T=9\). On the other hand, the SwinIR shallow feature extraction module only uses a convolution layer to extract features. Having the shallow feature map, \(\mathbf{\mathcal{X}}_0\), the deep feature extraction can be done, i.e.,
\begin{equation}
\mathbf{\mathcal{X}}_{\textit{DF}} = F_{\textit{DF}}(\mathbf{\mathcal{X}}_0),
\end{equation} 
where \(F_{\textit{DF}}(\cdot)\) contains \(NB\) number of MTSBlocks for MTSNet in contrast to RSTBs in SwinIR.
  
Figure \ref{fig:MTSBlock} illustrates the basic block of SwinIR, where MSA refers to the multi-head self-attention operator. MSA operations are non-linear, whereas the proposed MTS is a linear operation. To introduce non-linearity in the absence of activation functions, a multi-head-gate attention (MHG) module is proposed in this study. The architecture and design details of the MHG module will be explained in the next subsection.

In the feed-forward module, SwinIR employs an MLP operator; in contrast, MTSNet utilizes the following DepthWise convolution Gate (DWG):  
\begin{equation}
    \text{DWG}(\mathbf{\mathcal{X}}) = \left( f_{3} \left( g_{3}(\mathbf{\mathcal{X}}) \right) \right) \odot \left( f_{4}(g_{4}(\mathbf{\mathcal{X}})) \right),
\end{equation}
where, \(g_{3}\) and \(g_{4}\) denotes standard \(1 \times 1\) convolutions, while \(f_{3}\) and \(f_{4}\) are \(3 \times 3\) depthwise convolutions, akin to the gate operation described in \cite{restormer}. In conclusion, MSA and MLP layers form the Swin Transformer Layer (STL), and \(L\) number of STLs in a sequential manner, followed by a \(3\times 3\) standard convolution layer form the Swin (Transformer) Block, RSTB. On the other hand, MTS is supported by MHG and DWG, which collectively form MTS Nonlinear Layer (MTSNL), and \(L\) number of MTSNLs followed by a \(3\times 3\) standard convolution layer make MTSBlock as depicted in Figure \ref{fig:MTSBlock}.  Layer Normalization \cite{ba2016layer} is applied at the beginning of each block and before the feed-forward module, as is done in the Swin blocks.  

\begin{figure*}[t]
\centering
\begin{minipage}[b]{0.28\linewidth}
  \centering
  \includegraphics[width=\linewidth]{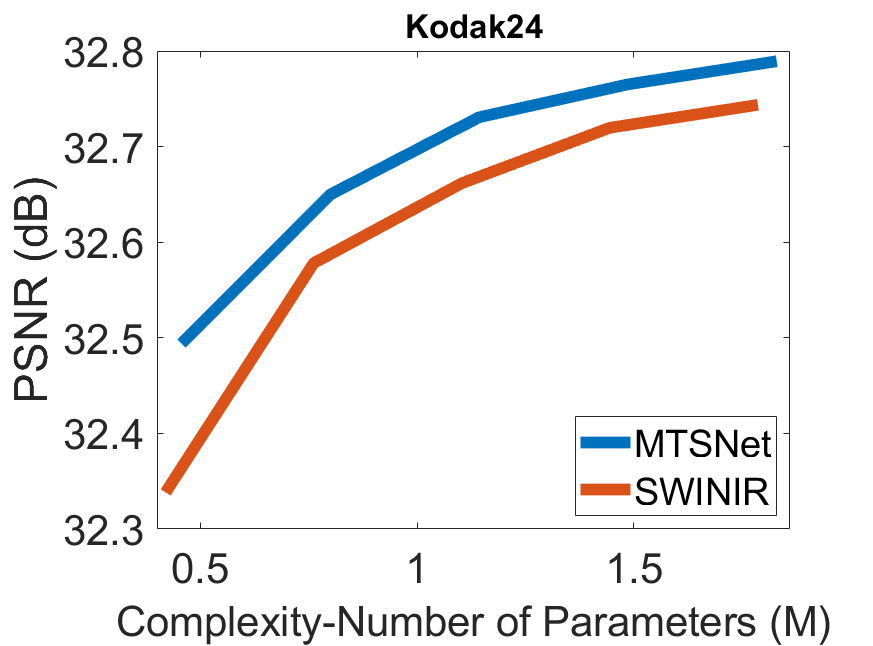}
\end{minipage}
\begin{minipage}[b]{0.28\linewidth}
  \centering
  \includegraphics[width=\linewidth]{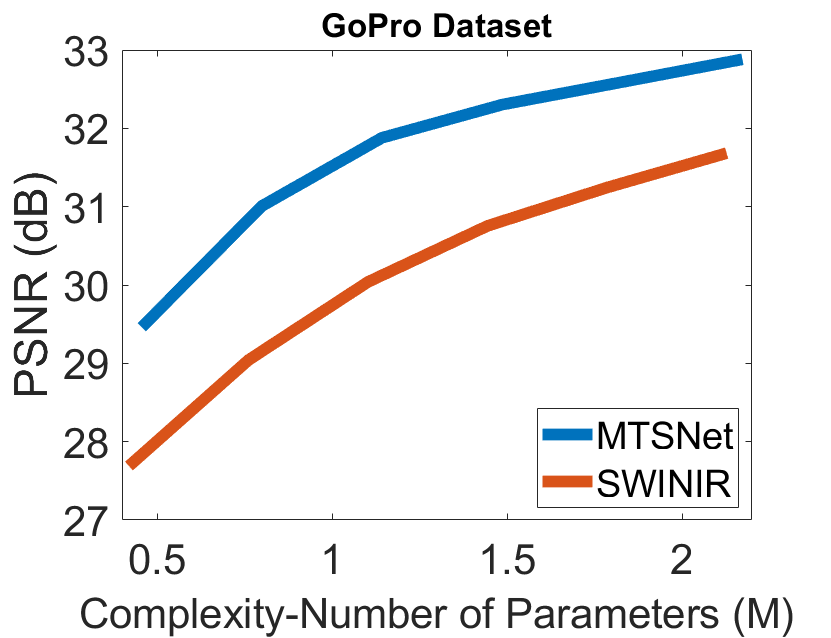}
\end{minipage}
\begin{minipage}[b]{0.28\linewidth}
  \centering
  \includegraphics[width=\linewidth]{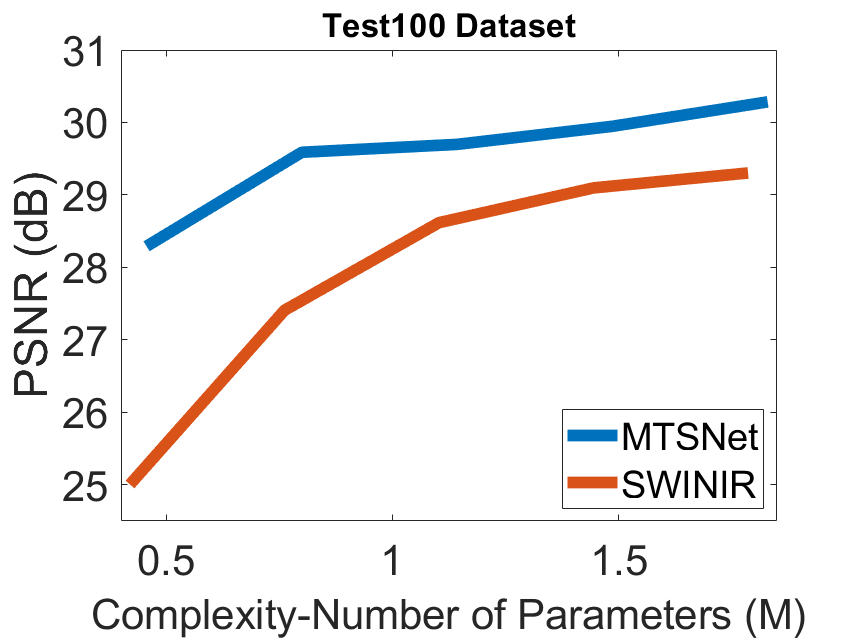}
\end{minipage}

\vspace{0.5ex}

\begin{minipage}[b]{0.28\linewidth}
  \centering
  \includegraphics[width=\linewidth]{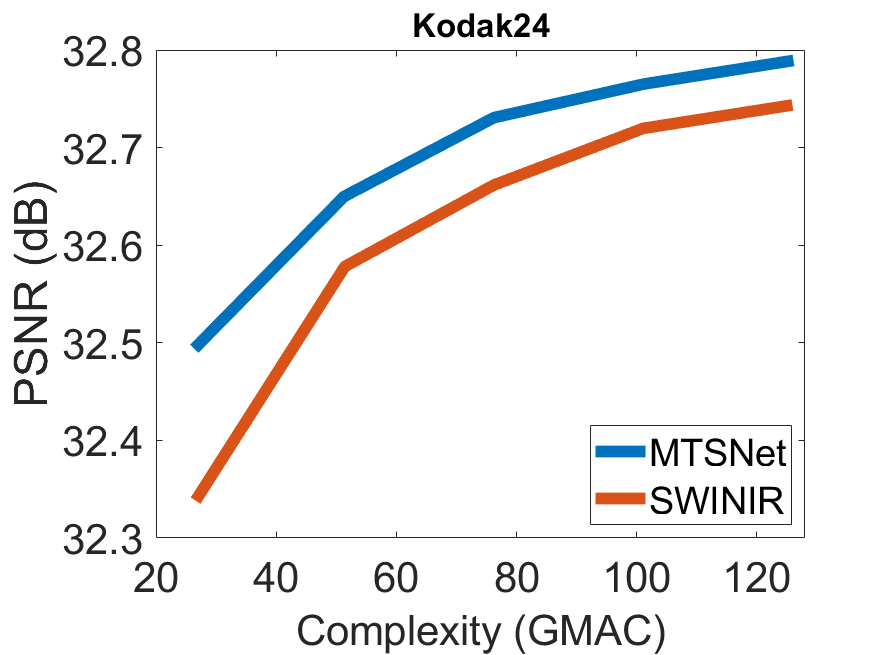}
  \vspace{0.5ex}
  \small Denoising (Kodak24)
\end{minipage}
\begin{minipage}[b]{0.28\linewidth}
  \centering
  \includegraphics[width=\linewidth]{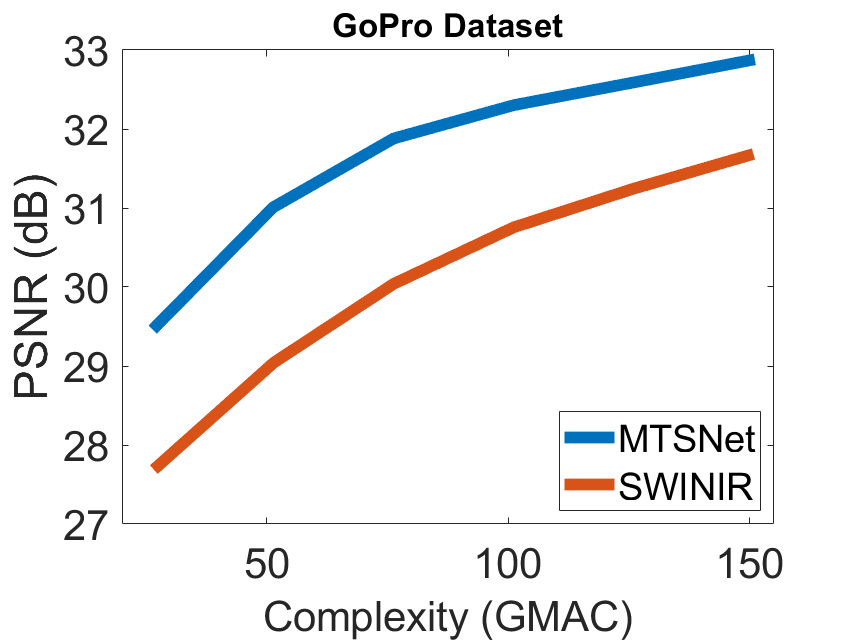}
  \vspace{0.5ex}
  \small Deblurring (GoPro)
\end{minipage}
\begin{minipage}[b]{0.28\linewidth}
  \centering
  \includegraphics[width=\linewidth]{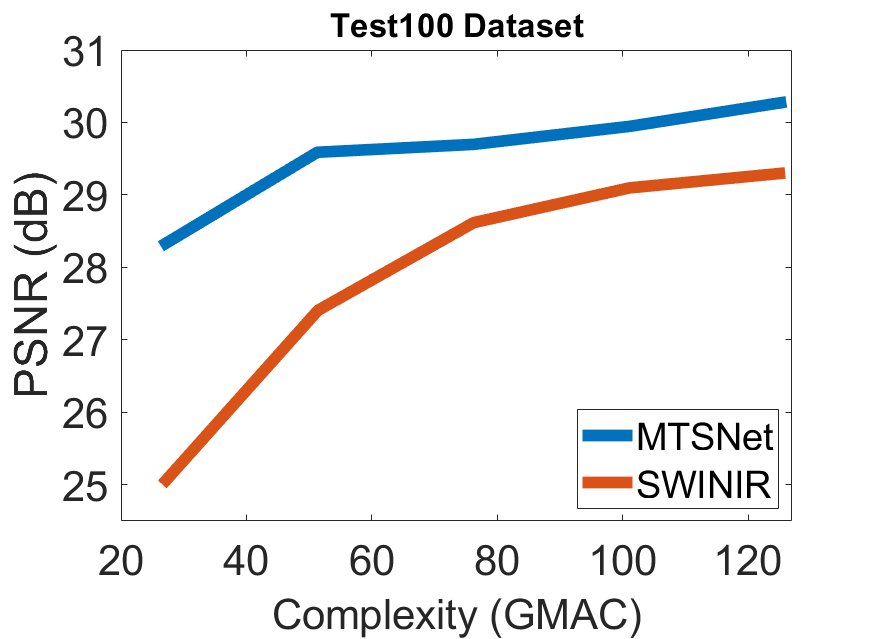}
  \vspace{0.5ex}
  \small Deraining (Test100)
\end{minipage}

\caption{Comparison of MTSNet and SwinIR with varying number of blocks for three image restoration tasks: denoising (Kodak24, \(\sigma=25\)), deblurring (GoPro), and deraining (Test100).}
\label{fig:denoising_MTSNetSwinIR}
\end{figure*}

Then, both SwinIR and MTSNet architectures consist of a global residual connection and a predefined number of their own corresponding blocks. In MTSNet, the residual connection is applied to the output of the high-quality (HQ) image recovery module, i.e.,
\begin{equation}
    \widehat{\mathbf{\mathcal{S}}} = F_{\textit{HQ}}(F_{\textit{DF}}(\mathbf{\mathcal{X}}_0)) + \mathbf{\mathcal{S}}_{\textit{LQ}},
\end{equation}
where the HQ image recovery module maps \(H \times W \times C\) feature maps to a \(H \times W \times 3\) residual signal via a standard convolution layer, i.e., \(F_{\textit{HQ}}(\cdot) = \text{Conv}_{3 \times 3}(\cdot)\), and \(\widehat{\mathbf{\mathcal{S}}} \in \mathbb{R}^{H \times W \times 3}\) denotes the recovered image.

\begin{figure}[]
\centering
  \includegraphics[width=1.05\linewidth]{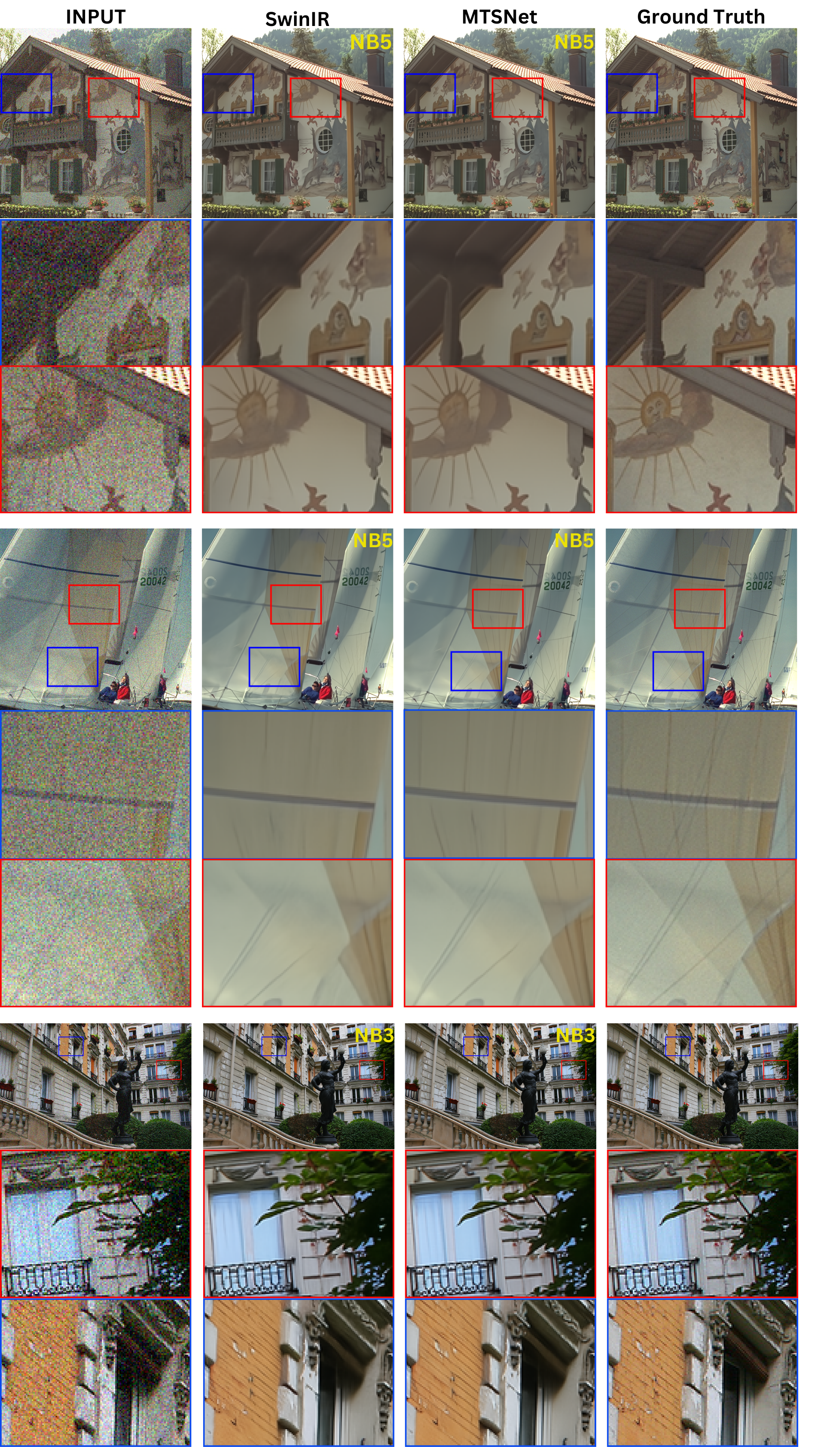}

  \caption{Visual quality comparison of MTSNet and SwinIR with varying
number of blocks for image denoising task on Kodak 24 and Urban 100 datasets. }
\label{fig:noise_swin_visual}
\end{figure}

\subsubsection{Activation-Free Non-linearity via Multi-Head Gate (MHG)}
During the past few years, there has been a noticeable trend within the machine learning community toward the adoption of \textit{gated operations}, which are increasingly being used to replace traditional non-linear activation functions, such as Rectified Linear Units (ReLU)~\cite{relu}. These gated operations typically involve element-wise multiplication between the outputs of a linear transformation and a linear transformation followed by a non-linear activation. A typical example of this is the Gated Linear Unit (GLU)~\cite{GLN}, defined as: \(
\text{Gate}(\mathbf{\mathcal{X}}, g_1, g_2, \sigma) = g_1(\mathbf{\mathcal{X})} \odot \sigma(g_2(\mathbf{\mathcal{X}})),
\) where \(g_1\) and \(g_2\) are linear operators, \(\sigma\) is a non-linear activation such as the sigmoid, and \(\odot\) denotes element-wise multiplication. The integration of these gated operations improves the representational capacity of neural networks by introducing non-linearity into their intermediate outputs. Many different versions of gated operations also exist in literature, including the Gaussian Error Linear Unit \cite{gelu} (GELU) and the simplified gate operation~\cite{Nafnet}. The latter, defined as \(\text{SGate}(\mathbf{\mathcal{X}}) = \text{Conv}_{1 \times 1}\left(\mathcal{X} \right ) \odot \text{Conv}_{1 \times 1}\left(\mathcal{X} \right ) \), has also been shown to bring significant efficiency in computer vision tasks. 

Inspired by this promising foundation, in this study, we introduce the following multi-head-gate attention (MHG) module:
\begin{equation}
    \text{MHG}(\mathbf{\mathcal{X}}, g_1, g_2) = \sum_{i=1}^{H} \left( f_{2i} \left( g_{1i}(\mathbf{\mathcal{X}}) \right) \right) \odot \left( f_{2i}(g_{2i}(\mathbf{\mathcal{X}})) \right)
\end{equation}
where \(g_{ji}\) are realized by group wise \( 1\times 1\) convolution, \(1 \times 1 ~ gconv\), \(f_{ji}\), are depthwise \( 3\times 3 \) convolutions, \(H\) is the number of heads, and \(\odot\) is the element wise multiplication operation. Group-wise convolution divides the feature map along the channel dimension, effectively reducing computational complexity while preserving a close connection among the channels. In addition, applying the operation and summing, i.e., multi-head gate operation, enhances the network’s non-linear representation capacity. 

\subsubsection{MTSNet vs SwinIR Implementation Details}
Both the MTSNet and SwinIR architectures have been implemented and trained using varying numbers of their respective block components. Specifically, the RSTB (the basic SwinIR block) and MTSBlock both consist of four layers, denoted as \(L=4\), with the MTSBlock containing \(C=56\) channels and the SwinIR containing \(C=90\) channels, respectively. This design choice has been made to ensure a comparable number of parameters and a similar level of computational complexity between the two competing architectures. For MTSBlock, the window sizes are selected as \(w=[8,16,32,64] \) for the first layer and progressively decrease to \(w = [8]\) in the last layer of each block. In MTSBlock's MHG operator, the number of heads is selected as \(H=4\), and the feed-forward expansion factor is set to \(2.2\). The total number of tensor summations is set to \(T=3\) for all the MTSBlocks and \(T=9\) for the shallow feature extraction module. In the SwinIR architecture, the number of heads and feed-forward expansion factor (MLP ratio) is set to 6 and 2, respectively, as specified in the original SwinIR paper. 

\begin{figure}[]
\centering
  \hspace{-0.2cm}
  \includegraphics[width=0.99\linewidth]{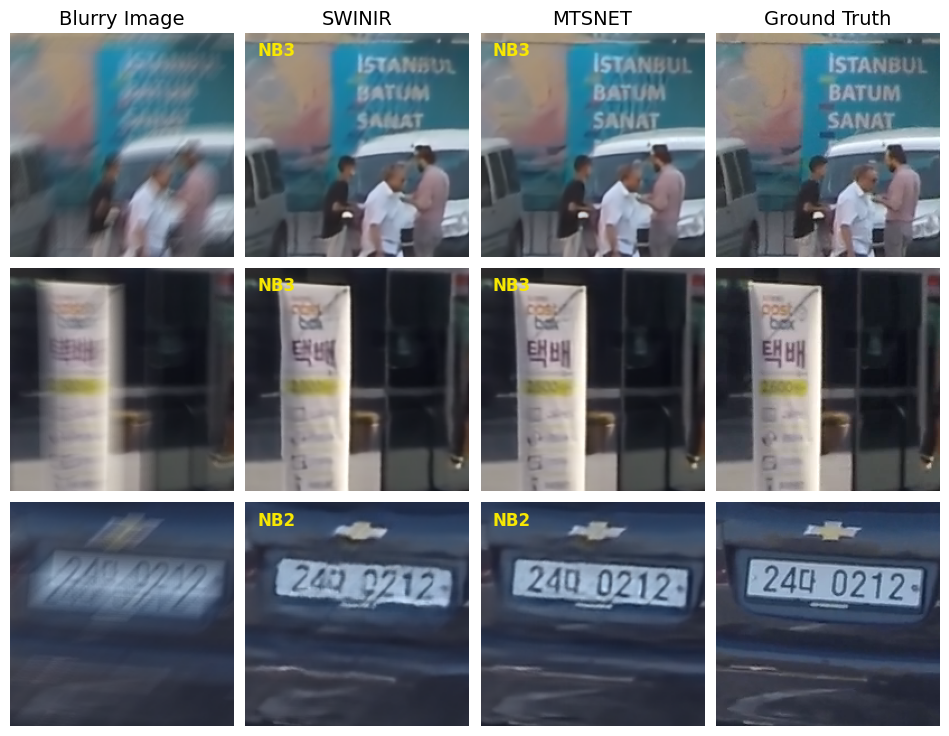}

  \caption{Visual quality comparison of MTSNet and SwinIR with varying
block counts for the image deblurring task in the GoPro Dataset }
\label{fig:deblurring_MTSNetSwinIR_visual}
\end{figure}

\begin{figure}[]
\centering
  \hspace{-0.2cm}
  \includegraphics[width=0.99\linewidth]{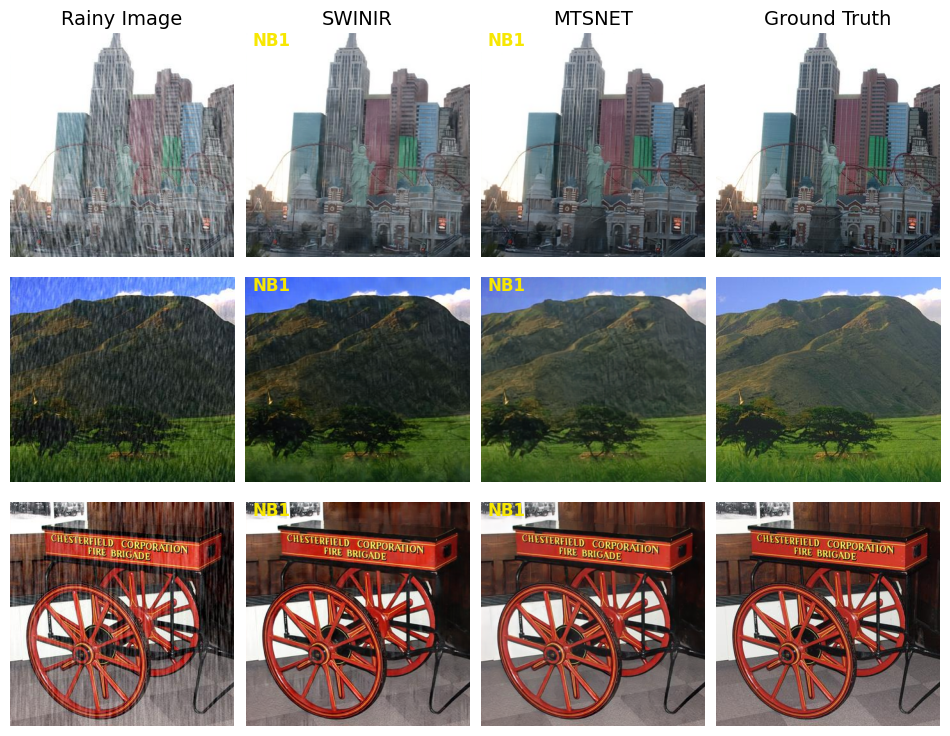}

  \caption{Visual quality comparison of MTSNet and SwinIR with varying
block counts for image deraining task in the Test100 Dataset. }
\label{fig:deraining_MTSNetSwinIR_visual}
\end{figure}

\subsubsection{MTSNet vs SwinIR Comparison}
In order to train the models for different image restoration tasks, namely image denoising, deblurring, and deraining, the same training datasets presented in Section \ref{convolutionvsMTS} were used i.e., WED, DIV2K, Flickr2K, BSD500 for denoising (considering different noise levels \(\sigma = 15, 25, 50\)), the GoPro training dataset for the deblurring task, and Rain13K for deraining. The models were trained with AdamW~\cite{AdamW} optimizer ($\beta_1 = 0.9$, $\beta_2 = 0.999$, weight decay = 0), and L1 loss was used for all models. A cosine annealing~\cite{loshchilov2016sgdr} scheduler was applied, with a starting learning rate of \( lr = 1\text{e}^{-3} \) for MTSNet models. SwinIR models, on the other hand, were trained with a lower learning rate of \( lr = 2\text{e}^{-4} \), as they were observed to be unstable at higher learning rates. Progressive learning~\cite{restormer} strategy is applied with starting path-size \(256 \times 256\) and progressively increased to \(512 \times 512\) for image deblurring, and to \(384 \times 384\) for image deraining and image denoising. 

\begin{figure*}[]
\centering
  \includegraphics[width=0.9\linewidth]{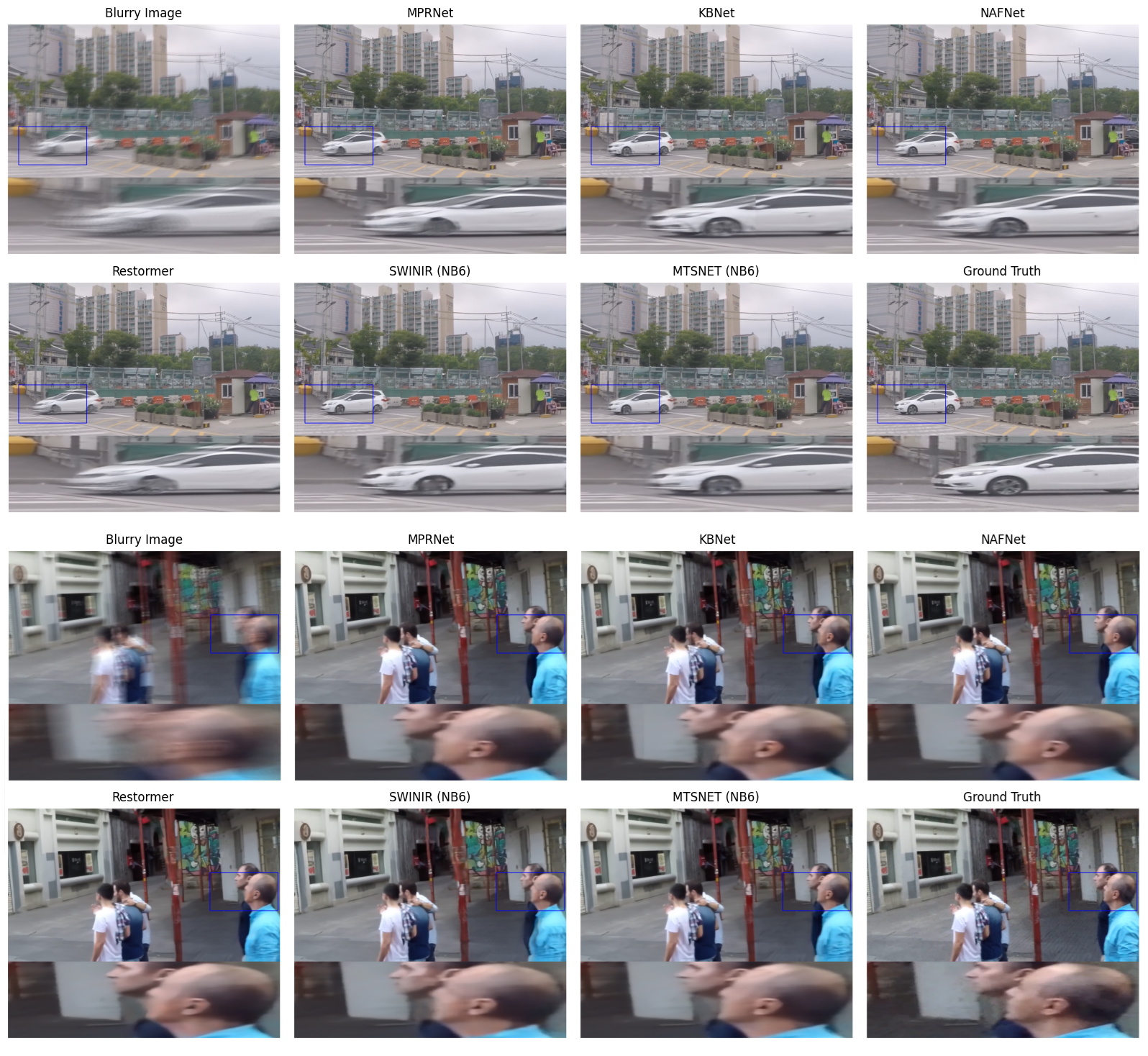}
  \caption{Visual comparison of outputs from various image deblurring methods on the GoPro Dataset.}
\label{fig:deblur_visual}
\end{figure*}

To ensure a fair comparison, we conducted training on various versions of 
SwinIR and MTSNet with increasing depth by systematically varying the total number of blocks for each network. Specifically, we examined models with block numbers ranging from 1 to 5 for the image denoising and deraining tasks and from 1 to 6 for the image deblurring task. As shown in Figure \ref{fig:denoising_MTSNetSwinIR}, MTSNet demonstrates a significant performance gain when compared to its SwinIR counterpart of similar complexity. Visual comparison of the outputs is also reported in Figures \ref{fig:noise_swin_visual}, \ref{fig:deblurring_MTSNetSwinIR_visual}, and \ref{fig:deraining_MTSNetSwinIR_visual} for the denoising, deblurring, and deraining tasks. The results demonstrate a significant improvement in edge preservation in the outputs of MTSNets compared to those obtained from SwinIR networks under different configurations. 
 
\subsection{Comparison with SoTa Image Restoration Networks}
In order to evaluate where the proposed MTSNet stands in terms of the performance–complexity trade-off among SoTa image restoration networks, we conducted an extensive analysis comparing it with various leading image restoration networks. First, different-depth versions of MTSNet were compared with SoTa image denoising networks. Table \ref{Table-I} reports the comparison of five variants of MTSNet, each constructed with a different number of blocks, against SoTa image denoising networks across four different datasets and three different noise levels (\(\sigma= 15, 25, 50\)). While MTSNet achieves comparable performance to SoTa networks in terms of reconstruction quality (in PSNR), it is particularly notable for its efficiency with respect to the number of learnable parameters, demonstrating a favorable performance–complexity trade-off.

Second, for image deblurring tasks, we compare our MTSNets with SoTa image deblurring networks: DeepDeblur~\cite{nah2017deep},  SRN~\cite{tao2018scale}, DGN~\cite{DGN}, PSS-NSC~\cite{gao2019dynamic}, MT-RNN~\cite{park2020multi}, DMPHN~\cite{zhang2019deep}, PVDNet~\cite{PVDNet}, BANET~\cite{banet}, MPRNET~\cite{MPRNET}, HINet~\cite{chen2021hinet}, Restormer~\cite{restormer}, NAFNet~\cite{Nafnet} and KBNet~\cite{zhang2023kbnet}. Figure \ref{fig:deblurring_all} reports the PSNR (in dB) versus the number of trainable parameters for each network. Although MTSNets are highly compact in terms of the number of learnable parameters, they achieve PSNR performance comparable to the deeper SoTa counterparts. In Figure \ref{fig:deblur_visual}, the outputs of SoTa networks are visually compared with those of MTSNet. In these challenging cases, where images have high levels of blurring and require a large receptive field, MTSNet is observed to produce visually superior results compared to SoTa networks.
\begin{figure}[h]
\centering
  \includegraphics[width=0.99\linewidth]{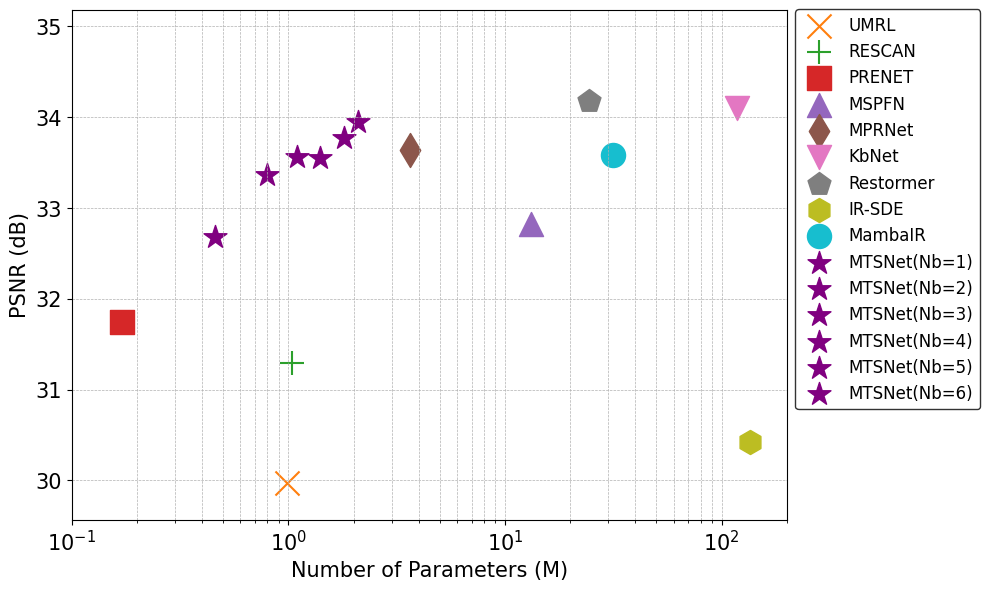}
  \caption{Deraining results of competing networks on the Test2800 dataset. }
\label{fig:deraining_alls}
\end{figure}

\begin{figure}[h]
\centering
  \includegraphics[width=0.99\linewidth]{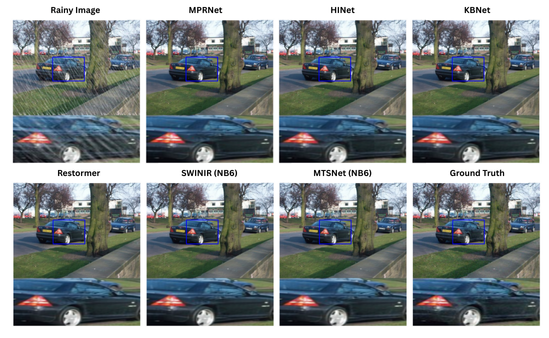}
  \caption{Visual comparison of the deraining performance on the Test2800 dataset. }
\label{fig:deraining_visual}
\end{figure}

Finally, we compared MTSNets with various leading single image deraining networks: UMRL~\cite{yasarla2019uncertainty}, RESCAN~\cite{li2018recurrent}, PreNet~\cite{ren2019progressive}, MSPFN~\cite{mspfn2020}, MPRNet~\cite{MPRNET}, KBNet~\cite{zhang2023kbnet}, Restormer~\cite{restormer}, IR-SDE~\cite{luo2023image}, MambaIR~\cite{guo2024mambair}. Figure \ref{fig:deraining_alls} reports the deraining results in terms of PSNR versus the number of learnable parameters. In Figure \ref{fig:deraining_visual}, the outputs of MTSNet are also compared visually. Once again, despite having a lower number of parameters, MTSNet is observed to produce outputs comparable to those of SoTa networks.

\section{Discussions}

\begin{table}[h] \scriptsize
\caption{Performance comparison (PSNR in dB) of MTSNet vs SwinIR. The results are on the color image denoising task with \(\sigma=25\) on Kodak 24 dataset -- means training is unstable due to large learning rates. }
\setlength\tabcolsep{2.6pt}
\centering
\label{stability}
\begin{tabular}{|l|lllll|}
\hline
                          & \multicolumn{5}{l|}{\begin{tabular}[c]{@{}l@{}}Number of \\ Blocks\end{tabular}}                                          \\ \hline
Network and Learning Rate & \multicolumn{1}{l|}{NB=1}  & \multicolumn{1}{l|}{NB=2}  & \multicolumn{1}{l|}{NB=3}  & \multicolumn{1}{l|}{NB=4}  & NB=5  \\ \hline
SwinIR $lr=10^{-3}$       & \multicolumn{1}{l|}{32.35} & \multicolumn{1}{l|}{32.59} & \multicolumn{1}{l|}{-}     & \multicolumn{1}{l|}{-}     & -     \\ \hline
SwinIR $lr=10^{-4}$       & \multicolumn{1}{l|}{32.35} & \multicolumn{1}{l|}{32.59} & \multicolumn{1}{l|}{32.66} & \multicolumn{1}{l|}{32.72} & 32.74 \\ \hline
MTSNet $lr= 10^{-3}$      & \multicolumn{1}{l|}{32.49} & \multicolumn{1}{l|}{32.65} & \multicolumn{1}{l|}{32.73} & \multicolumn{1}{l|}{32.77} & 32.79 \\ \hline
\end{tabular}
\end{table}

\begin{table}[h]\scriptsize
\caption{Comparison of architectures on STL10 and MNIST with classification accuracy and parameter count. TensorSum+ uses a standard linear final layer due to multiscale constraints.}

\centering
\setlength\tabcolsep{2.6pt}
\centering
\label{stability_tensorsum}
\begin{tabular}{|l|l|c|c|c|}
\hline
\textbf{Architecture} & \textbf{Dataset} & \textbf{Window Size} & \textbf{Accuracy (\%)} & \textbf{\#Parameters} \\
\hline
TensorSum            & STL10 & 4$\times$8 & 36.175  & 4265     \\
TTNet                & STL10 & -          & 38.9    & 528948   \\
TensorSum+           & STL10 & 4$\times$8 & 46.0125 & 277380   \\
Linear (MLP)         & STL10 & -          & 42.2125 & 764716042 \\
\hline
TensorSum            & MNIST & 4$\times$8 & 97.37   & 2709     \\
TTNet                & MNIST & -          & 97.25   & 61614    \\
TensorSum+           & MNIST & 4$\times$8 & 97.74   & 11384    \\
Linear (MLP)         & MNIST & -          & 98.22   & 1059850  \\
\hline
\end{tabular}
\end{table}

Our investigation has revealed that, in comparison to dense layers within compact architectures, MTS layers demonstrate considerably enhanced stability throughout the training process as discussed in Section \ref{denselayervsmts}. In this section, we also mentioned that the last layers of the classifier networks
can not be implemented in a patch-wise manner, but can be done via GTS. Alternatively, this layer can still be replaced with a standard MLP layer, which yields TensorSum+, and the results can be further improved as reported in Table \ref{stability_tensorsum}. Moreover, MTSNets exhibit superior stability compared to transformer models. Specifically, when evaluating MTSNets against one of the SoTa transformer architectures in the image restoration task, SwinIR, we observe that SwinIR faces significant challenges in convergence with higher learning rates across diverse depth lengths for a variety of image restoration tasks, as reported in Table \ref{stability}. In contrast, MTSNets can be effectively trained using larger learning rates across a broad spectrum of depth configurations.  This finding suggests that increasing the size of the receptive field while preserving the stability of the training substantially improves overall performance.

\section{Conclusions}
\label{conclusion}
In this paper, we introduced a tensorial and learnable factorization, MTS, as a novel neural network operator that achieves very large receptive fields, even in the first layer, therefore enhancing efficiency in deep learning architectures. In this new operator, the receptive field size is adjustable with pre-defined window sizes. This novel neural network operator was first tested with small-scale image classifiers and compression models, and we showed that it is much more stable and delivers high performance compared to MLP layers. We then extended our experiments to compare the new operator with convolutional layers. Convolutional layers can also be written as a special factorized form of dense layers in the form of matrix-vector multiplication. Due to their high stability, convolutions form the foundation of deep neural networks.
In many image restoration experiments, the proposed new operator showed an obvious performance improvement over convolutional layers.
Next, we extended our comparison to the more recent backbone operators, i.e., transformer blocks that consist of self-attention and feedforward modules. When we modified the basic building blocks of transformers to the one we created and, consisting of the proposed MTS and a non-linear operator, MHG, we demonstrated the advantages of the proposed block in terms of both complexity and performance across various image restoration tasks. In this study, we did not focus on optimizing network architectures; instead, we simply stacked blocks, similar to the vanilla transformer structure of SwinIR. In future work, more complex network architectures can be developed using the proposed backbone operator, such as various UNet structures or hybrid models.
Even in this proof-of-concept vanilla structure, MTSNet achieved performance values comparable to state-of-the-art image restoration networks with significantly less number of learnable parameters. As a result, the experimental validations show that the proposed MTS is a promising backbone neural layer, and we will further investigate its usage in different tasks and neural structures.

\ifCLASSOPTIONcaptionsoff
  \newpage
\fi

{\small
\bibliographystyle{IEEEtran}
\bibliography{egbib}
}

\end{document}